\documentclass[a4paper]{article}\usepackage[]{graphicx}\usepackage[]{color}
\makeatletter
\def\maxwidth{ %
  \ifdim\Gin@nat@width>\linewidth
    \linewidth
  \else
    \Gin@nat@width
  \fi
}
\makeatother

\definecolor{fgcolor}{rgb}{0.345, 0.345, 0.345}

\usepackage{framed}
\makeatletter
 {\par\unskip\endMakeFramed%
 \at@end@of@kframe}
\makeatother

\definecolor{shadecolor}{rgb}{.97, .97, .97}
\definecolor{messagecolor}{rgb}{0, 0, 0}
\definecolor{warningcolor}{rgb}{1, 0, 1}
\definecolor{errorcolor}{rgb}{1, 0, 0}
\newenvironment{knitrout}{}{} % an empty environment to be redefined in TeX

\usepackage{alltt}
\usepackage{authblk} % author affiliation
\usepackage{amsmath} % align env.
\usepackage{amssymb} % expectation symbol
\usepackage{relsize} % for \mathlarger
\usepackage[comma]{natbib} % bibliography
\usepackage{hyperref} % url & citation links
\usepackage{url} % to handle url's in & *.bib
\usepackage{rotating} % sidewaystable
\usepackage[final]{changes} % manual change markup
\usepackage[noframe]{showframe} % to show margins
\usepackage{enumitem} % resuming enumerate
\allowdisplaybreaks
\usepackage{setspace}
\renewcommand{\a}{\alpha}
\newcommand{\A}[1][]{\mathbf{A}_{#1}}
\renewcommand{\b}{\beta}
\newcommand{\B}[1][]{\mathbf{B}_{#1}}
\newcommand{\C}[1][]{\mathbf{C}_{#1}}

\newcommand{\e}{\boldsymbol{\eta}}
\newcommand{\E}[1][]{\mathbf{E}_{#1}}
\newcommand{\p}[1][]{\boldsymbol{\phi}_{#1}}
\newcommand{\N}[1][]{\mathbf{N}_{#1}}
\renewcommand{\P}[1][]{\boldsymbol{\Phi}_{#1}}
\renewcommand{\t}[1][]{\boldsymbol{\theta}_{#1}}
\newcommand{\T}[1][]{\mathbf{\Theta}_{#1}}
\newcommand{\V}[1][]{\mathbf{V}_{#1}}
\newcommand{\x}{\boldsymbol{x}}
\newcommand{\W}[1][]{\mathbf{W}_{#1}}
\newcommand{\Y}[1][]{\mathbf{Y}_{#1}}
\newcommand{\Z}[1][]{\mathbf{Z}_{#1}}
\newcommand{\Prod}[2][]{\mathlarger\prod_{#2}^{#1}} % 2 args., 1st optional w/ empty default
\newcommand{\one}[1]{\boldsymbol{1}\left\{#1\right\}}
\IfFileExists{upquote.sty}{\usepackage{upquote}}{}
\begin{document}

\title{
  \textbf{A Review of Stochastic Block Models and Extensions for Graph Clustering}
}
\author[1]{Clement Lee}
\author[2,3]{Darren J Wilkinson}
\affil[1]{Department of Mathematics and Statistics, Lancaster University, UK}
\affil[2]{School of Mathematics, Statistics and Physics, Newcastle University, UK}
\affil[3]{The Alan Turing Institute, UK}
\affil[ ]{\texttt{clement.lee@lancaster.ac.uk}}
\maketitle

\begin{abstract}
  There have been rapid developments in model-based clustering of graphs, also known as block modelling, over the last ten years or so. We review different approaches and extensions proposed for different aspects in this area, such as the type of the graph, the clustering approach, the inference approach, and whether the number of groups is selected or estimated. We also review models that combine block modelling with topic modelling and/or longitudinal modelling, regarding how these models deal with multiple types of data. How different approaches cope with various issues will be summarised and compared, to facilitate the demand of practitioners for a concise overview of the current status of these areas of literature.  
\end{abstract}

\section{Introduction} \label{sect.intro}

Stochastic block models (SBMs) are an increasingly popular class of models in statistical analysis of graphs or networks. They can be used to discover or understand the (latent) structure of a network, as well as for clustering purposes. We introduce them by considering the example in Figure \ref{fig.example}, in which the network consists of 90 nodes and 1192 edges. The nodes are divided into 3 groups, with groups 1, 2 and 3 containing 25, 30 and 35 nodes, respectively. The nodes within the same group are more closely connected to each other, than with nodes in another group. Moreover, the connectivity pattern is rather ``uniform''. For example, compared to nodes 2 to 25, node 1 does not seem a lot more connected to other nodes, both within the same group or with another group. In fact, this model is generated by taking each pair of nodes at a time, and simulating an (undirected) edge between them. The probability of having such an edge or not is independent of that of any other pair of nodes. For two nodes in the same group, that is, of the same colour, the probability of an edge is 0.8, while for two nodes in different groups, the edge probability is 0.05.

\begin{knitrout}
\definecolor{shadecolor}{rgb}{0.969, 0.969, 0.969}\color{fgcolor}\begin{figure}[h!]

{\centering \includegraphics[width=0.95\linewidth]{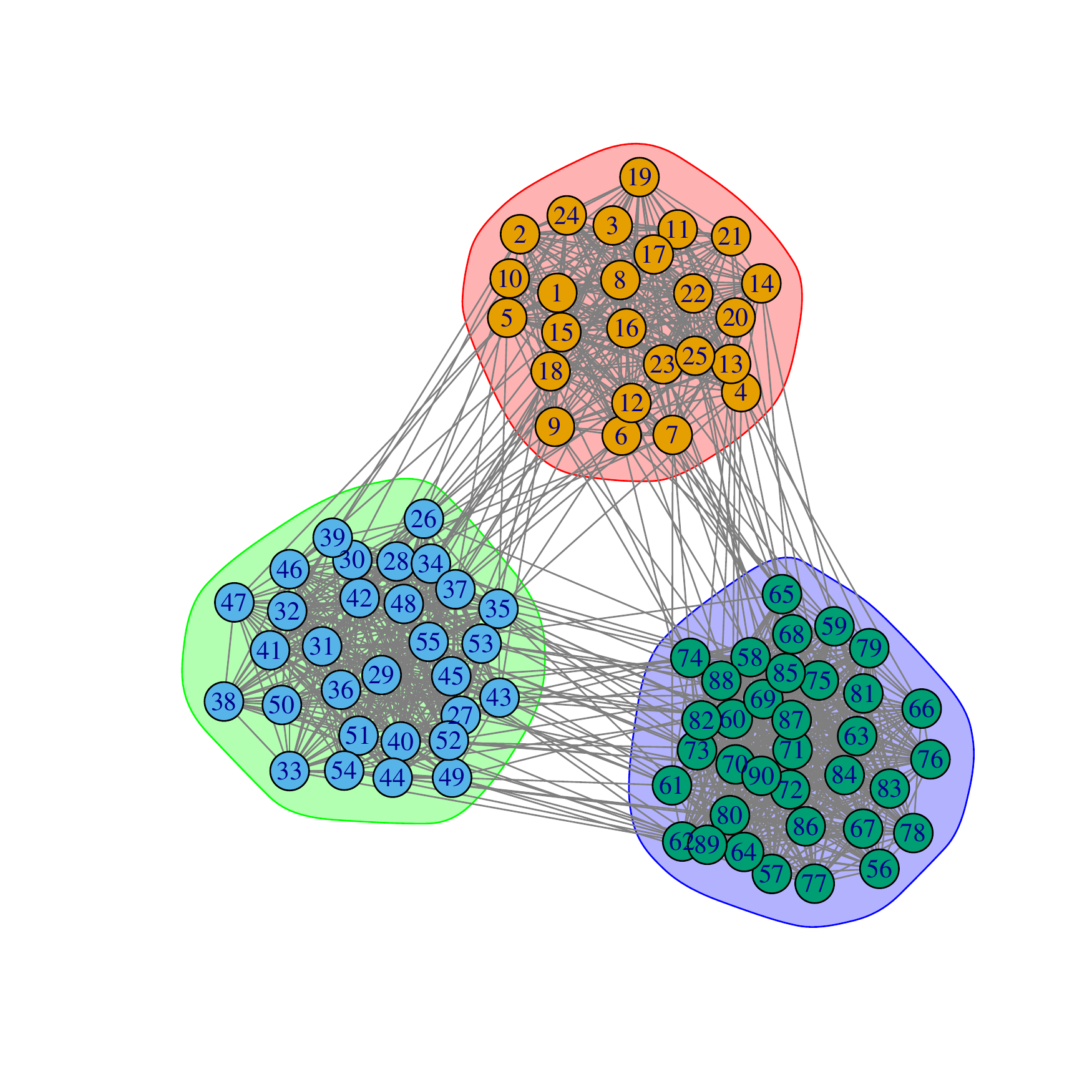} 

}

\caption[Network of 90 nodes]{Network of 90 nodes.}\label{fig.example}
\end{figure}

\end{knitrout}

The above example is a \textit{simulation} from a simple SBM, in which there are two essential components, both of which will be explained in Section \ref{sect.graph}. The first component is the vector of group memberships, given by $(\underbrace{1~1\cdots1}_{25}\underbrace{2~2\cdots2}_{30}\underbrace{3~3\cdots3}_{35})^T$ in this example. The second is the block matrix, each element of which represents the edge probability of two nodes, \textit{conditional on their group memberships}. According the description of the generation of edges above, the block matrix is
\begin{align}
  \left(\begin{array}{lll}
    0.8 & 0.05 & 0.05 \\
    0.05 & 0.8 & 0.05 \\
    0.05 & 0.05 & 0.8
  \end{array}\right).\label{eqn.example_block}
\end{align}

Given the group memberships, the block matrix, and the assumptions of an SBM (to be detailed in Section \ref{sect.graph}), it is straightforward to generate a synthetic network for \textit{simulation} purposes, as has been done in the example. It will also be straightforward to evaluate the likelihood of data observed, for \textit{modelling} purposes. However, in applications to real data, neither the group memberships nor the block matrix is observed or given. Therefore, the goal of fitting an SBM to a graph is to infer these two components simultaneously. Subsequently, the usual statistical challenges arise:

\begin{enumerate}
\item \label{q.modelling} Modelling: How should the SBM be structured or extended to realistically describe real-world networks, with or without additional information on the nodes or the edges?
\item \label{q.inference} Inference: Once the likelihood can be computed, how should we infer the group memberships and the block matrix? Are there efficient and scalable inference algorithms?
\item \label{q.selection} Selection and diagnostics: Can we compute measures, such as Bayesian information criterion (BIC) and marginal likelihood, to quantify and compare the goodness of fit of different SBMs?
\end{enumerate}

Inferring the group memberships is essentially clustering the nodes into different groups, and the number of groups, denoted by $K$, is quite often unknown prior to modelling and inference. This brings about another challenge:

\begin{enumerate}[resume]
\item \label{q.K} Should we incorporate $K$ as a parameter in the model, and infer it in the inference? Or should we fit an SBM with different fixed $K$'s, and view finding the optimal $K$ as a model selection problem?
\end{enumerate}

In this article, we will review the developments of SBMs in the literature, and compare how different models deal with various issues related to the above questions. Such issues include the type of the graph, the clustering approach, the inference approach, and selecting or estimating the (optimal) number of groups. The issue with the number of groups $K$ is singled out because how differently it is related to questions 1-3 depends largely on the specific model reviewed.

Quite often, other types of data, such as textual and/or temporal, appear alongside network data. One famous example is the Enron Corpus, a large database of over 0.6 million emails by 158 employees of the Enron Corporation before the collapse of the company in 2001. While the employees and the email exchanges represent the nodes and the edges of the email network, respectively, the contents and creation times of the emails provide textual and dynamic information, respectively. It has been studied by numerous articles in the literature, including \cite{zmlgz06,mwc07,pdbe08,fsx09,xfs10,gmgfb12,scfs12,dbs13,xh13,mrv15,blz16,cblr18}. Another example is collections of academic articles, in which the network, temporal and textual data come from the references/citations between the articles, their publication years, and their actual contents, respectively. In these cases, further questions can be asked:

\begin{enumerate}[resume]
\item \label{q.extension} Can longitudinal and/or textual modelling, especially for cluster purposes, be incorporated into the SBMs, to utilise all information available in the data?
\item \label{q.extension_aspects} How are inference, model selection/diagnostics, and the issue with $K$ dealt with under the more complex models?
\end{enumerate}

A relevant field to answering question \ref{q.extension} is topic modelling, in which the word frequencies of a collection of texts/articles are analysed, with the goal of clustering the articles into various topics. While topic modelling and SBMs are applicable to different types of information, textual for the former and relational for the latter, their ultimate goals are the same, which is model-based clustering of non-numerical data. Similar issues to the aforementioned ones for SBMs are also dealt with in various works, which will therefore be reviewed in this article.

While incorporating longitudinal and/or topic modelling into SBMs are possible, it should be noted that they are well-developed and large fields on their own. Due to the scope of this article, we will focus on works that are more recent or relatively straightforward extensions of SBMs to these two fields. We hope, from these reviewed interdisciplinary works, to provide future directions to a comprehensive model that handles multiple types of information simultaeously and deals with the questions raised satisfactorily.

While articles on SBMs form a major body of works in the literature, especially over the last decade or so, they should not be completely separated from other methods or algorithms in statistical network analysis. For example, community detection methods and latent space models are two highly related fields to SBMs, and their developments are interwined with each other. Therefore, several important works, such as reviews of these two topics or articles which make connections with SBMs, will be mentioned as well, for the sake of comprehensiveness.

The rest of this article is organised as follows. A simple version of the SBM is introduced in Section \ref{sect.graph}. Extensions of the SBM regarding the type of graph are reviewed in Section \ref{sect.extension}. Models that relax the usual clustering approach, in which each node is assumed a single group, are introduced in \ref{sect.graph_cluster}. Related models and methods for graphs to the SBM are discussed in Section \ref{sect.related}. The inference approaches and the related issue of the number of groups are discussed in Sections \ref{sect.graph_inf} and \ref{sect.graph_K}, respectively. Models which incorporate longitudinal modelling are \ref{sect.graph_longitudinal}. Topic modelling is briefly introduced in Section \ref{sect.topic}, and its incorporation in SBMs is reviewed in \ref{sect.both}. A summary and comparison of models are provided in Section \ref{sect.compare_all}, and the discussion in Section \ref{sect.discussion} concludes the article.

\section{Stochastic block models} \label{sect.graph}
In this section, we shall first formulate a basic version of the stochastic block model (SBM) and mention the concept of stochastic equivalence, illustrated by continuing with the example in Section \ref{sect.intro}. This will pave the way for Section \ref{sect.extension}, where we consider different extensions to accommodate additional information about the graph, to better describe real-world networks, and to potentially lead to more scalable inference algorithms.

To introduce the terminology, we consider a graph $\mathcal{G}=(\mathcal{N},\mathcal{E})$, where $\mathcal{N}$ is the node set of size $n:=|\mathcal{N}|$, and $\mathcal{E}$ is the edge list of size $M:=|\mathcal{E}|$. In the example in Figure \ref{fig.example}, $\mathcal{N}=\{1,2,\ldots,90\}$, $n=90$, and $M=1192$. We call a pair of nodes a \textit{dyad}, and consider the existence or absence of an edge for the dyad $(p,q)$, through the use of the $n\times{}n$ adjacency matrix, denoted by $\Y$, which is another useful representation of the graph. If $\mathcal{G}$ is undirected, as is the example, $\Y[pq]=\Y[qp]=1$ if $p$ and $q$ have an edge between them, 0 otherwise, where $\mathbf{M}_{rs}$ represents the $(r,s)$-th element of matrix $\mathbf{M}$. By construction, under an undirected graph, $\Y$ is symmetric along the major diagonal. If $\mathcal{G}$ is directed, $\Y[pq]=1~(0)$ represents an edge (non-edge) from $p$ \textit{to} $q$, and is independent of $\Y[qp]$, for $\Y$ need not be symmetric. We assume $\Y[pp]=0$, that is, no node has a \textit{self-edge}, as this is quite often, but not always, assumed in the models reviewed. The adjacency matrix for the example is shown in Figure \ref{fig.example_adj}, where black and white represent 1 and 0, respectively. Graphs with binary adjacency matrices are called binary graphs hereafter.

\begin{knitrout}
\definecolor{shadecolor}{rgb}{0.969, 0.969, 0.969}\color{fgcolor}\begin{figure}[h!]

{\centering \includegraphics[width=0.55\linewidth]{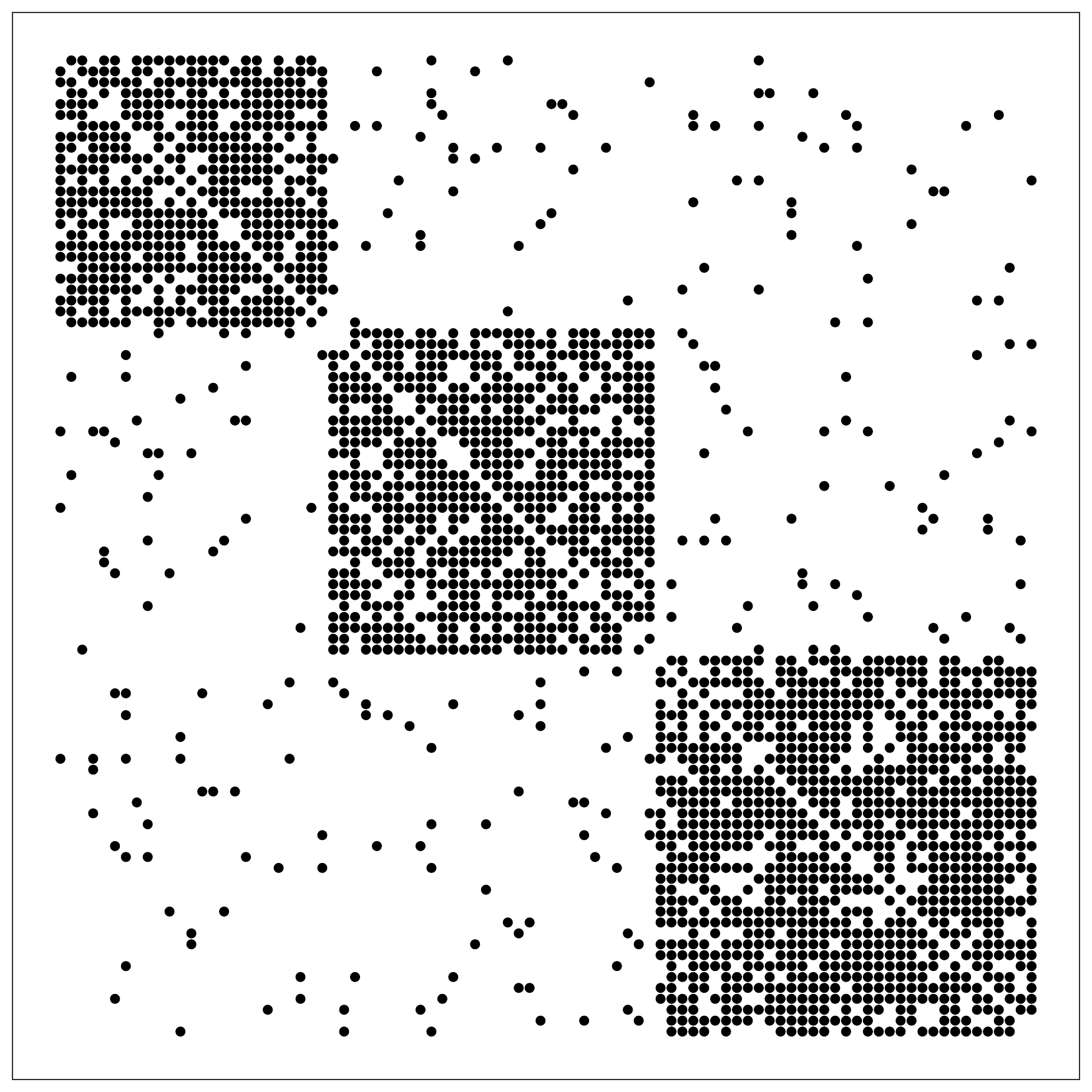} 

}

\caption{Adjacency matrix of example in Figure \ref{fig.example}.}\label{fig.example_adj}
\end{figure}

\end{knitrout}

In the SBM, each node belongs to one of the $K(<n)$ groups, where $K=3$ in the example. As the groups are unknown before modelling, for node $p=1,2,\ldots,n$ also defined is a $K$-vector $\Z[p]$, all elements of which are 0, except exactly one that takes the value 1 and represents the group node $p$ belongs to. For instance, as nodes 1, 45 and 90 in the example belong to groups 1, 2 and 3, respectively, we have $\Z[1]=(1~0~0)^T$, $\Z[45]=(0~1~0)^T$, and $\Z[90]=(0~0~1)^T$. Also defined is an $n\times{}K$ matrix $\Z:=\left(\Z[1]~\Z[2]~\cdots~\Z[n]\right)^T$, such that $\Z[pi]$ is the $i$-th element of $\Z[p]$.

The group sizes can be derived from $\Z$, and are denoted by $\N=(\N[1]~\N[2]~\cdots~\N[K])^T$. Essentially, $\N[i]$ is the sum, or number of non-zero elements, of the $i$-th column of $\Z$. In the example, $\N[1]=25$, $\N[2]=30$, $\N[3]=35$. Finally, the $K\times{}K$ edge matrix between \textit{groups} can be derived from $\Z$ and $\Y$. It is denoted by $\E$, where $\E[ij]$ represents the number of edges between groups $i$ and $j$ in undirected graphs, and from group $i$ to group $j$ in directed ones. In the example, $\E$ is symmetric as $\mathcal{G}$ is undirected, and $\E[11]=245$, $\E[22]=341$, $\E[33]=481$, $\E[12]=\E[21]=37$, $\E[23]=\E[32]=52$, and $\E[31]=\E[13]=36$.

In order to describe the generation of the edges of $\mathcal{G}$ according to the groups the nodes belong to, a $K\times{}K$ block matrix, denoted by $\mathbf{C}$, is introduced. If $\mathcal{G}$ is undirected, for $1\leq{}i\leq{}j\leq{}K$, $\C[ij]\in[0,1]$ and represents the probability of occurrence of an edge between a node in group $i$ and a node in group $j$. Here $\C$ is symmetric, as is \eqref{eqn.example_block} in the example. If $\mathcal{G}$ is directed, for $1\leq{}i,j\leq{}K$, $\C[ij]$ represents the probability of occurrence of a directed edge from a node in group $i$ to a node in group $j$, and $\C$ needs not be symmetric. Note that no rows or columns in $\C$ need to sum to 1.

Whether $\mathcal{G}$ is undirected or directed, the idea of the block matrix $\C$ means that the dyads are conditionally independent given the group memberships $\Z$. In other words, $\Y[pq]$ follows the Bernoulli distribution with success probability $\Z[p]^T\C\Z[q]$, and is independent of $\Y[rs]$ for $(p,q)\neq(r,s)$, \textit{given $\Z[p]$ and $\Z[q]$}. This implies that the total number of edges between any two blocks $i$ and $j$ is a Binomial distributed random variable with mean equal to the product of $\C[ij]$ and the number of dyads available. For undirected and directed graphs, the latter term is $\N[i]\N[j]/2$ and $\N[i]\N[j]$, respectively. In fact, Figure \ref{fig.example_adj} can be viewed as a realisation of simulating from the Binomial distribution with the respective means. Conversely, the densities of each pair of blocks in the adjacency matrix, calculated to be
\begin{align}
  \left(\begin{array}{lll}
    0.817 & 0.049 & 0.041 \\
    0.049 & 0.784 & 0.05 \\
    0.041 & 0.05 & 0.808 
  \end{array}\right),\nonumber
\end{align}
are, as expected, close to \eqref{eqn.example_block}.

\subsection{Stochastic equivalence} \label{sect.sto_equ}
The assumption that the edge probability of a dyad depends solely on their memberships (and $\C$) is based on the concept of \textit{stochastic equivalence} (see, for example, \cite{ns01}). In less technical terms, for nodes $p$ and $q$ in the same group, $p$ has the same (and independent) probability of connecting with node $r$, as $q$ does. This echoes the observation in the example that node 1 does not seem more connected to the whole network than other nodes in the same group are. While such probability depends on the group membership of $r$, the equality still holds between $p$ and $q$.

\subsubsection{Block modelling and community detection}
The concept of stochastic equivalence in itself does not require that nodes in the same group are more connected within themselves, than with nodes in another group. Essentially, the elements along the major diagonal of $\C$ are not necessarily higher than the off-diagonal elements. However, this phenomenon, which is also observed in the example, is often the goal of community detection, a closely related topic to SBMs. Therefore, it is sometimes taken into account in the modelling in the \textit{assortative} or \textit{affinity} SBM by, for example, \cite{gmgfb12} and \cite{law16}. Community detection algorithms and assortative SBMs will be further discussed in Sections \ref{sect.com_det} and \ref{sect.assort}, respectively.

\subsection{Modelling and likelihood} \label{sect.model_lik}
Given $\Z$ and $\C$, we can write down the likelihood based on the assumption of edges being Bernoulli distributed conditional on the group memberships. If $\mathcal{G}$ is undirected, again assuming no self-edges, the likelihood is
\begin{align}
  &\quad\pi\left(\Y|\Z,\C\right)=\Prod[n]{p<q}\pi(\Y[pq]|\Z,\C)\nonumber\\
  =&~\Prod[n]{p<q}\left[\left(\Z[p]^T\C\Z[q]\right)^{\Y[pq]}\left(1-\Z[p]^T\C\Z[q]\right)^{\left(1-\Y[pq]\right)}\right].\label{eqn.graph_lik_hard}
\end{align}
If $\mathcal{G}$ is directed, we replace the index in the product in \eqref{eqn.graph_lik_hard} from $p<q$ to $p\neq{}q$. The model with this likelihood will be called the Bernoulli SBM hereafter. With a change of index, \eqref{eqn.graph_lik_hard} can be written as
\begin{align}
  &\quad\pi(\Y|\Z,\C)=\Prod[K]{i\leq{}j}\C[ij]^{\E[ij]}(1-\C[ij])^{\N[ij]-\E[ij]},\label{eqn.graph_lik_hard_re}
\end{align}
where $\N[ij]=\N[i]\N[j]/2$ if $i\neq{}j$, $\N[ij]=\N[i](\N[i]-1)/2$ if $i=j$. Multiplying over the group indices $i,j$ will become more useful than over the node indices $p,q$ in some cases.

As mentioned in Section \ref{sect.intro}, when applying SBMs to real-world data, usually neither $\Z$ nor $\C$ is known, and has to be inferred. Therefore, assumptions have to be made before modelling and inference. For $p=1,2,\ldots,n$, we assume that the latent variable $\Z[p]$ is independent of $\Z[q]$ \textit{apriori}. Also, we assume that $\Pr(\Z[pi]=1)=\theta_{i}$, where $\theta_{i}$ is the $i$-th element of the $K$-vector $\t=\left(\theta_1~\theta_2~\ldots~\theta_K\right)^T$ such that $\sum_{i=1}^K\theta_i=1$. Essentially, the latent group $\Z[p]$ follows the multinomial distribution with probabilities $\t$, which means
\begin{align}
  \pi\left(\Z|\t\right)=\Prod[n]{p=1}\Z[p]^T\t=\Prod[n]{p=1}\t^T\Z[p]=\Prod[K]{i=1}\theta_i^{\N[i]}.\label{eqn.graph_z_hard}
\end{align}
%where $m_i=\sum_{p=1}^n\one{\Z[p]=i}$, and $\one{E}$ is the indicator function of event $E$. 
A further assumption can be made that $\t$ arises from the Dirichlet$(\left(\a\boldsymbol{1}_K\right)$ distribution, of which the parameter $\a$ comes from a Gamma$(a,b)$ prior. This will be useful in Section \ref{sect.poisson}. We shall defer actual inference to Section \ref{sect.graph_inf}, and first examine the extensions and variants of SBM in the following section.

\section{Type of graph and extensions of the SBM} \label{sect.extension}
In this section, we briefly revisit the lineage of the SBM, discuss how it is extended for binary graphs, and introduce models for valued graphs, to answer question \ref{q.modelling} in Section \ref{sect.intro}. Special attention will be paid to two increasingly useful variants: the degree-corrected and the microcanonical SBMs.

SBMs originated from their deterministic counterparts. \cite{bba75} illustrated an algorithm to essentially permute the rows and columns of the adjacency matrix $\Y$. The rearranged adjacency matrix contains some submatrices with zeros only, some others with at least some ones. The former and latter kinds of submatrices are summarised by 0 and 1, respectively, in what they called the ``blockmodel'', which can be viewed as the predecessor of the block matrix $\C$. \cite{wbb76} followed this line but also calculated the densities of the blocks in some of their examples. The stochastic generalisation of the ``blockmodel'' was formalised by \cite{hll83}. While \cite{ww87} applied the SBM to real directed graphs, they assumed that the block structure is known \textit{a priori}. \cite{sn97} and \cite{ns01} studied \textit{a posteriori} blocking, meaning that the groups are initially unknown and to be inferred via proper statistical modelling, for 2 and an arbitrary number of groups, respectively. These lead to the Bernoulli SBM for binary graphs described in Section \ref{sect.graph}.

\subsection{Binary graphs}
Apart from \cite{sn97}, binary graphs have been studied by numerous models. In the absence of additional information, such as covariates or attributes associated with the edges or nodes, the focus of any developments has been on proposing alternative models, or optimising the number of groups $K$. For example, the mixed membership models in \cite{abfx08,fsx09,xfs10,fcx15} and \cite{law16} allowed each node to belong to multiple groups (Section \ref{sect.graph_cluster}). \cite{mgj09} and \cite{msh11} proposed latent feature models (Section \ref{sect.feature}) that deviate from the SBM, while \cite{zhou15} proposed an edge partition model (Section \ref{sect.related}). \cite{msh11} and \cite{fcx15} also, alongside \cite{kgbs13} and \cite{mmfh13}, \textit{modelled} $K$ in different ways (Section \ref{sect.graph_K_modelled}), whereas \cite{lba12,cl15} and \cite{yan16} derived various criteria for \textit{selecting} $K$ (Section \ref{sect.criteria}). \cite{lw18} proposed a model for binary directed \textit{acyclic} graphs (DAGs), in which the possible combinations for $(\Y[pq],\Y[qp])$ are $\{(0,0),(0,1),(1,0)\}$, and no directed cycles of any length are allowed. This means that if we go from a node to an arbitrary neighbour along the direction of the edges and perform this ``random walk'' recursively, it is not possible to reach the original node.

Quite often temporal data are avaialble to the networks observed. Therefore, there are numerous models for binary graphs that incoporate longitudinal modelling, including \cite{fsx09,xfs10,yczgj11,xh13,ty11,ty14,fcx15,mrv15}, and \cite{len17}. Therefore, they will be discussed separately in Section \ref{sect.graph_longitudinal}.

While longitudinal SBMs deal with multiple layers of graphs (in a temporal order) directly, there are cases in which the multiple layers of graphs are aggregated, and the observed graph contains one layer only, with information about the specific layers lost. \cite{vmgs16} proposed a multilayer SBM for such a situation. Using the example of 2 layers, they assumed the adjacency matrix $\Y^l$ of layer $l=1,2$ is generated by a Bernoulli SBM according to block matrix $\C^l$ independently, and then considered two versions of the model. In the first version, there is an edge between $p$ and $q$ in the graph to be observed if an edge for the dyad exists in \textit{either} layer, which means $\Y[pq]=1-(1-\Y[pq]^1)(1-\Y[pq]^2)$. In the second, the existence of the edge in the observed graph requires the existence of the corresponding edge in both layers, which menas $\Y[pq]=\Y[pq]^1\Y[pq]^2$. \cite{vmgs16} found that the second version is more predictive (than the usual single-layer SBM) when it comes to link prediction or network reconstruction (Section \ref{sect.graph_inf_pred}) for real-world networks, which may therefore be better described as an aggregation of multiple layers.

\subsection{Valued graphs}
The Bernoulli SBM can be easily generalised or modified for valued graphs. Apart from the aforementioned undirected and directed graphs, \cite{ns01} also studied directed signed graphs, where $\Y[pq]$ can take the value $-1, 0$, or $1$. If we consider the dyad of a node of group $i$ and a node of group $j$, and further define two $K\times{}K$ matrices $\mathbf{D}$ and $\mathbf{E}$ such that $\mathbf{D}_{ij}$ and $\mathbf{E}_{ij}$ represent the probabilities that the dyad takes the value $0$ and $-1$, respectively, subject to $\C[ij]+\mathbf{D}_{ij}+\mathbf{E}_{ij}=1$, the edge probability, with a slight abuse of notation, is
\begin{align}
  \Pr(\Y[pq]=1|\Z,\C,\mathbf{D},\mathbf{E})=\Z[p]^T\left(\C^{\one{\Y[pq]=1}}\mathbf{D}^{\one{\Y[pq]=0}}\mathbf{E}^{\one{\Y[pq]=-1}}\right)\Z[q],\nonumber%\label{eqn.graph_ypq_ns01}
\end{align}
where $\one{A}$ is the indicator function of event $A$. They also studied graphs for tournaments, which means the edge between nodes $p$ and $q$ can represent a match between the two nodes. The possible combinations for $(\Y[pq],\Y[qp])$ are $\{(-1,1),(1,-1),(0,0)\}$, corresponding to a loss, win and tie for $p$, respectively. The specification of the edge probability is omitted here because of the higher \textit{notational} complexity. There is limited adoption of models in the literature for this kind of graphs.

\cite{kks06} considered a graph in which multiple edges between two nodes can be accounted for. Instead of taking the product over all dyads, they assumed each edge arises from the usual Bernoulli$(\Z[p]^T\C\Z[q])$ distribution \textit{after} drawing the two nodes $p$ and $q$ independently from a multinomial distribution with weights $\boldsymbol{\phi}=(\phi_1~\phi_2\cdots\phi_n)^T$. If we assume the $m$-th edge corresponds to the dyad $(p_m,q_m)$, the likelihood is
\begin{align}
  \pi(\mathcal{E}|\boldsymbol{\phi},\Z,\C)=\Prod[M]{m=1}\left[\phi_{p_m}\phi_{q_m}\left(\Z[{p_m}]^T\C\Z[{q_m}]\right)^{\Y[{p_mq_m}]}\left(1-\Z[{p_m}]^T\C\Z[{q_m}]\right)^{1-\Y[{p_mq_m}]}\right].\nonumber
\end{align}
While \cite{kks06} mainly considered bipartite graphs, we consider the two types of nodes as essentially the same set of nodes, to align with the notation in other models. 

\cite{yczgj11} mainly worked with binary undirected graphs in their dynamic SBM (see Section \ref{sect.graph_longitudinal}), but briefly extended to the valued version, where $\Y[pq]$, now the discrete number of interactions for dyad $(p,q)$, is being modelled by a geometric distribution:
\begin{align}
  \pi\left(\Y[pq]|\Z,\C\right)=\left(\Z[p]^T\C\Z[q]\right)^{\Y[pq]}\left(1-\Z[p]^T\C\Z[q]\right).\label{eqn.graph_ypq_yczgj11}
\end{align}
\cite{dbs13} also proposed a model based on the SBMs for network dynamics. In its simplest version, for the dyad $(p,q)$, the interactions are being modelled by a Poisson process with intensity $\exp\left(\Z[p]^T\C\Z[q]\right)$, where the elements of $\C$ are not bounded by $0$ and $1$ here. Further modifications of the model are discussed in Section \ref{sect.graph_longitudinal}.

\subsection{Poisson and degree-corrected SBMs} \label{sect.poisson}
\cite{kn11} also worked with (undirected) valued graphs, but arguably in a more natural way. They first redefined $\Y[pq]$ to be the number of edges for the dyad $(p,q)$ following a Poisson distribution, and $\C[ij]$ the \textit{expected} number of edges from a node in group $i$ to a node in group $j$. The density of $\Y[pq]$ is now
\begin{align}
  \pi\left(\Y[pq]|\Z,\C\right)=\left(\Y[pq]!\right)^{-1}\exp\left(-\Z[p]^T\C\Z[q]\right)\left(\Z[p]^T\C\Z[q]\right)^{\Y[pq]}.\label{eqn.graph_ypq_poisson}
\end{align}
They argued that, in the limit of a large sparse graph where the edge probability equals the expected number of edges, this version of the SBM, called the Poisson SBM, is asymptotically equivalent to the Bernoulli counterpart in \eqref{eqn.graph_lik_hard}. To further modify the model, a parameter $\phi_p$ is introduced for each node, subject to a constraint $\sum_{p=1}^{n}\phi_p\one{\Z[pi]=1}=1$ for every group $i$, so that the expected number of edges for the dyad $(p,q)$ is now $\phi_p\phi_q\Z[p]^T\C\Z[q]$. The density of $\Y[pq]$ becomes
\begin{align}
  \pi\left(\Y[pq]|\Z,\C,\p\right)=\left(\Y[pq]!\right)^{-1}\exp\left(-\phi_p\phi_q\Z[p]^T\C\Z[q]\right)\left(\phi_p\phi_q\Z[p]^T\C\Z[q]\right)^{\Y[pq]},\label{eqn.graph_ypq_kn11}
\end{align}
where $\p=(\phi_1~\phi_2~\cdots~\phi_n)^T$. This is termed the degree-corrected (DC) SBM. The parameters $\phi_p$ and $\C[ij]$ have natural interpretations as their maximum likelihood estimates (MLEs) are the ratio of $p$'s degree to the sum of degrees in $p$'s group, and the total number of edges between groups $i$ and $j$, respectively. While \cite{kn11} have also considered self-edges in their model, such treatment is omitted here for easier notational alignment. 

Of importance is the reason behind the DC-SBM. \cite{kn11} argued that the then existing SBMs usually ignores the variation in the node degrees in real-world networks. This is quite evident as, under the original SBM, the expected degree is the same for all nodes in each group, given $\Z$ and $\C$. \cite{kn11} illustrated that the DC-SBM managed to discover the known factions in a karate club network, while the original counterpart failed to do so. \replaced{Quite a few recent works built upon the DC-SBM in various ways. For instance, }{}\cite{ysjkmzzz14} provided an approach to model selection (Section \ref{sect.criteria}) between the two versions of SBMs.\replaced{ Also, \cite{ls19b} introduced additional parameters to $\p$ as a means of taking the \textit{assortativeness} into account; see Section \ref{sect.assort}.}{}

\cite{mmfh13} managed to integrate out the model parameters of the Poisson SBM in \eqref{eqn.graph_ypq_poisson}, to form a \textit{collapsed} SBM. Apart from \eqref{eqn.graph_lik_hard}, \eqref{eqn.graph_z_hard} and the Dirichlet distribution for $\t$, they also assumed \textit{apriori} each $\C[ij]$ is independent and identically distributed (i.i.d.) according to a Gamma$(\gamma,\beta)$ distribution (with mean $\gamma/\beta$). With a change of indices similar to \eqref{eqn.graph_lik_hard_re}, the likelihood can be written as
\begin{align}
  &\quad\pi(\Y|\Z,\C)=\Prod[n]{p<q}\left(\Y[pq]!\right)^{-1}\exp\left(-\Z[p]^T\C\Z[q]\right)\left(\Z[p]^T\C\Z[q]\right)^{\Y[pq]}\nonumber\\
  =&~\Prod[K]{i\leq{}j}\exp\left(-\C[ij]\N[i]\N[j]\right){\C[ij]}^{\E[ij]}\times\Prod[n]{p<q}(\Y[pq]!)^{-1}\label{eqn.graph_ypq_poisson_ij}
\end{align}
Now, the joint density of $\Y$ and $\Z$, with $\C$ and $\t$ integrated out, can be derived:
\begin{align}
  &\quad\pi(\Y,\Z|\a,\b,\gamma)=\pi(\Y|\Z,\beta,\gamma)\times\pi(\Z|\alpha)\nonumber\\
  =&~\int\pi(\Y|\Z,\C)\pi(\C|\b,\gamma)d\C\times\int\pi(\Z|\t)\pi(\t|\a)d\t\nonumber\\
  =&~\Prod[n]{p<q}(\Y[pq]!)^{-1}\times\mathlarger\int\left[\Prod[K]{i\leq{}j}e^{-\C[ij]\N[i]\N[j]}{\C[ij]}^{\E[ij]}\times\Prod[K]{i\leq{}j}e^{-\b\C[ij]}{\C[ij]}^{\gamma-1}\b^{\gamma}/\Gamma(\gamma)\right]d\C\nonumber\\
  &\quad\times\mathlarger\int\Prod[K]{i=1}\theta_i^{\N[i]}\times\left[\Gamma\left(K\a\right)\one{\sum_{i=1}^K\theta_i=1}\Prod[K]{i=1}\frac{\theta_i^{\a-1}}{\Gamma\left(\a\right)}\right]d\t\nonumber\\
  =&~\Prod[n]{p<q}(\Y[pq]!)^{-1}\times\Prod[K]{i\leq{}j}\frac{\Gamma(\gamma+\E[ij])\b^{\gamma}}{\Gamma(\gamma)\left(\b+\N[i]\N[j]\right)^{\gamma+\E[ij]}}\times\frac{\Gamma(K\a)\Prod[K]{i=1}\Gamma(\alpha+\N[i])}{\Gamma(\alpha)^K\Gamma(K\alpha+n)}\label{eqn.graph_ypq_mmfh13}
\end{align}
Now \eqref{eqn.graph_ypq_mmfh13} is a function of $\Y$, $\Z$ (through $\E$ and $\N$), and the three scalar (hyper-)parameters only. This becomes particularly useful in inference (Section \ref{sect.graph_inf_collapsed}). Also, \eqref{eqn.graph_ypq_mmfh13} can be compared with \eqref{eqn.graph_ypq_micro} in the microcanonical SBM below. Furthermore, if $\alpha$, $\beta$ and $\gamma$ are fixed or integrated out, $\pi(\Y,\Z)$ is the exponential of the \textit{integrated complete data log-likelihood} (ICL). It is a useful quantity when it comes to inference (Section \ref{sect.graph_inf_variational}) and model selection (Section \ref{sect.criteria}).

\cite{ajc15} introduced a unifying framework for modelling binary and valued graphs, by observing that both the Bernoulli and Poisson distributions belong to the exponential family of distributions. For example, \eqref{eqn.graph_lik_hard} can be written as
\begin{align}
  &\quad\pi\left(\Y|\Z,\C\right)=\Prod[n]{p<q}\left[\left(\Z[p]^T\C\Z[q]\right)^{\Y[pq]}\left(1-\Z[p]^T\C\Z[q]\right)^{\left(1-\Y[pq]\right)}\right]\nonumber\\
  =&~\exp\left\{\mathlarger\sum_{p<q}^n\left[\Y[pq]\log\left(\frac{\Z[p]^T\C\Z[q]}{1-\Z[p]^T\C\Z[q]}\right)+\log\left(1-\Z[p]^T\C\Z[q]\right)\right]\right\}\nonumber\\
  =&~\exp\left\{\mathlarger\sum_{p<q}^n\boldsymbol{S}(\Y[pq])^T\e(\Z[p]^T\C\Z[q])\right\},\label{eqn.graph_ypq_wsbm}
\end{align}
where $\boldsymbol{S}(x)=(x,1)^T$ and $\e=\left(\log\left(x/(1-x)\right),\log\left(1-x\right)\right)^T$ are both vector-valued functions, of the sufficient statistics and the natural parameters of the Bernoulli distribution, respectively. Now, with a different type of edges observed, $\boldsymbol{S}(x)$ and $\e(x)$ can be specified according to the appropriate distribution in the exponential family. In the case of the Poisson SBM \eqref{eqn.graph_ypq_poisson}, $\boldsymbol{S}(x)=(x,-\log(x!),-1)^T$ and $\e(x)=(\log{}x,1,x)^T$.

\cite{ajc15} also used this \textit{weighted} SBM to clarify the meaning of zeros in valued graphs, as they could mean a non-edge, an edge with weight zero, or missing data. To overcome this ambiguity, they extended \eqref{eqn.graph_ypq_wsbm}, so that the (log-)likelihood is a mixture of distributions:
\begin{align}
  \pi(\Y|\Z,\C)=\exp\left\{\psi\sum_{p<q}^n\boldsymbol{S}_1(\Y[pq])^T\e_1(\Z[p]^T\C\Z[q])+(1-\psi)\sum_{p<q}^n\boldsymbol{S}_2(\Y[pq])^T\e_2(\Z[p]^T\C\Z[q])\right\},\nonumber
\end{align}
where $\psi$ is the weight assigned to the two components. They can correspond to, for example, the Bernoulli and the Poisson SBM, for modelling the edge \textit{existence} and edge \textit{weight}, respectively.

\subsection{Microcanonical SBM} \label{sect.micro}
Arguably the most important recent developments is the microcanonical SBM and its nested version \citep{peixoto17a,peixoto17b}, as these works are a culmination of applying the principle of minimum description length (MDL) \citep{grunwald07,rb07}, a fundamental result regarding $K$ that is related to community detection \citep{peixoto13} (Section \ref{sect.com_det}), an efficient MCMC inference algorithm \citep{peixoto14a}, a hierarchical structure that models $K$ simultaneously \citep{peixoto14b} (Section \ref{sect.graph_K_modelled}), and an approach that models $\Z$ and $\C$ differently from \eqref{eqn.graph_z_hard}, leading to efficient inference algorithm (Section \ref{sect.graph_inf_monte_carlo}). The microcanonical SBM is mentioned here because it can be derived from modifying the Poisson SBM. The case for undirected graphs is illustrated here, with \eqref{eqn.graph_ypq_poisson_ij} utilised again. Next, for $1\leq{}i\leq{}j\leq{}K$ and \textit{conditional on $\Z$ and an extra parameter $\lambda$}, $\C[ij]$ is assumed to follow the Exponential distribution with rate $\N[i]\N[j]/\lambda$. This assumption replaces that according to \eqref{eqn.graph_z_hard}. By doing so, $\C$ can be integrated out from the product of \eqref{eqn.graph_ypq_poisson_ij} and the exponential density of $\C$:
\begin{align}
  &\quad\pi(\Y|\Z,\lambda)=\int\pi(\Y|\Z,\C)\pi(\C|\Z,\lambda)d\C\nonumber\\
  =&~\Prod[n]{p<q}(\Y[pq]!)^{-1}\times\mathlarger\int\left[\Prod[K]{i\leq{}j}e^{-\C[ij]\N[i]\N[j]}{\C[ij]}^{\E[ij]}\times\Prod[K]{i\leq{}j}e^{-\C[ij]\N[i]\N[j]/\lambda}\left(\N[i]\N[j]/\lambda\right)\right]d\C\nonumber\\
%  =&~\Prod{p<q}(\Y[pq]!)^{-1}\times\Prod{i\leq{}j}\left(\N[i]\N[j]/\lambda\right)\times\Prod{i\leq{}j}\int{}e^{-\C[ij]\N[i]\N[j](1+1/\lambda)}{\C[ij]}^{\E[ij]}d\C[ij]\nonumber\\
  =&~\Prod[n]{p<q}(\Y[pq]!)^{-1}\times\Prod[K]{i\leq{}j}\left(\N[i]\N[j]/\lambda\right)\times\Prod[K]{i\leq{}j}\frac{\Gamma(\E[ij]+1)}{\left[\N[i]\N[j](1+1/\lambda)\right]^{\E[ij]+1}}\nonumber\\
  =&~\Prod[n]{p<q}(\Y[pq]!)^{-1}\times\Prod[K]{i\leq{}j}\frac{\lambda^{\E[ij]}}{(1+\lambda)^{\E[ij]+1}}\times\Prod[K]{i\leq{}j}\frac{\E[ij]!}{(\N[i]\N[j])^{\E[ij]}}\nonumber\\
  =&~\underbrace{\frac{\lambda^M}{(1+\lambda)^{M+K(K+1)/2}}}_{\pi(\E|\Z,\lambda)}\times\underbrace{\Prod[n]{p<q}(\Y[pq]!)^{-1}\times\Prod[K]{i\leq{}j}\frac{\E[ij]!}{(\N[i]\N[j])^{\E[ij]}}}_{\pi(\Y|\E,\Z,\lambda)},\label{eqn.graph_ypq_micro}
\end{align}
where $M$, as defined in Section \ref{sect.graph}, is the total number of edges. Now, $\Z$ influences \eqref{eqn.graph_ypq_micro}, or what we call the \textit{integrated} likelihood, through $\E$ and $\N$ only. (We refrain from calling it the \textit{marginal} likelihood, which we refer to the likelihood with $\Z$ also integrated out.) Furthermore, this $\pi(\Y|\Z,\lambda)$ can be split into the product of the two underbraced terms, which is the joint likelihood of a microcanonical model \citep{peixoto17a}. It is termed ``microcanonical'' because of the hard constraints imposed, as $\Y$ and $\Z$ together fix the value of $\E$, and therefore
\begin{align}
  \pi(\Y|\Z,\lambda)=\pi(\Y,\E|\Z,\lambda)=\pi(\Y|\E,\Z,\lambda)\times\pi(\E|\Z,\lambda).\nonumber
\end{align}
Further marginalisation of $\lambda$, which is straightforward with one-dimensional integration of $\pi(\E|\Z,\lambda)$ in \eqref{eqn.graph_ypq_micro}, results in
\begin{align}
  \pi(\Y|\Z)=\pi(\Y|\E,\Z)\times\pi(\E|\Z),\label{eqn.graph_y_micro}
\end{align}
where no model parameters are involved, which however, as argued by \cite{peixoto17a}, is not compulsory in the microcanonical formulation. It is also called a nonparametric SBM, not because of having no parameters but because that $K$ is being modelled (Section \ref{sect.graph_K_modelled}). Finally, the same marginalisation of $\C$ (and $\lambda$) can be applied to \eqref{eqn.graph_ypq_kn11} to arrive at the microcanonical DC-SBM.

\subsection{Graphs with covariates or attributes}
\cite{tallberg05} proposed a model in which the group memberships $\Z$ depend on the covariates $\x$, through what they called a random utility model. Specifically, assume that $\x$ is $d$-dimensional, and associated with each group $i$ is a $d$-vector $\boldsymbol{\beta}_i$. The group membership of node $p$ is determind by $\Z[p]=\underset{i}{\text{argmax}}\left(\x_p^T\beta_i+\epsilon_{pi}\right)$, where $\epsilon_{pi}$ is an i.i.d. Gaussian-distruted error. In this way, the covariates $\x_p$ determine the memberships $\Z[p]$ through the group-specific vectors $\{\boldsymbol{\beta}_1,\boldsymbol{\beta}_2,\cdots,\boldsymbol{\beta}_K\}$ (and the error terms).

\cite{vhs13} also studied binary graphs as well as directed signed graphs. Furthermore, they proposed a model that connected the SBMs with exponential random graph models (ERGMs), another prominent class of social network analysis models which can be traced back to \cite{hl81}. In one example of the model by \cite{vhs13}, the edge probability is
\begin{align}
  \pi\left(\Y[pq]|\Z,\C,\x\right)\propto\exp\left[\psi{}f(\x,\Y[pq])+\left(\Z[p]^T\C\Z[q]\right)g(\x,\Y[pq])\right],\label{eqn.graph_ypq_vhs13}
\end{align}
where $\x$ are the covariates, and $f$ and $g$ are functions that may depend on $\x$. We do not specify the index of $\x$ because the covariates may depend on the nodes or the dyads. The parameter $\psi$ is constant to both $\x$ and the groups $p$ and $q$ belong to, while the use of $\C$, each element of which is possibly vector-valued and not bounded by $0$ and $1$, is to illustrate the dependence on the blocks and alignment with other models.

\cite{peixoto18a} extended the microcanonical SBM by incorporating attributes $\x$. Specifically, the joint density of $\Y$ and $\x$ is
\begin{align}
  \pi\left(\Y,\x|\Z\right)=\pi\left(\x|\Y,\Z\right)\pi\left(\Y|\Z\right),\label{eqn.graph_lik_peixoto18a}
\end{align}
where $\pi(\Y|\Z)$ is given by \eqref{eqn.graph_y_micro}, and the model is termed the nonparametric weighted SBM. Again, a hierarchical or nested structure can be incorporated, so that the \textit{groups} are modelled by another SBM, and so on, if necessary. Please see Section \ref{sect.graph_K} for further details.

\cite{sbknm19} proposed an \textit{attribute} SBM in which the group memberships $\Z$ of the nodes determine both the graph $\Y$ and the non-relational attributes $\x$, which are assumed conditionally independent given $\Z$. In the model formulation, in addition to a Bernoulli SBM for an undirected graph, they assumed that the attribute $\x[p]$ for node $p$ comes from a mixture of Gaussian distributions with weights $\t$, which are the probabilities for generating $\Z[p]$ as defined in Section \ref{sect.model_lik}. By utilising both $\x$ and $\Y$ together in inferring $\Z$, \cite{sbknm19} found that the two types of information can complement each other, and therefore their \textit{attribute} SBM is useful for link prediction (Section \ref{sect.graph_inf_pred}).

\section{Clustering approach}\label{sect.graph_cluster}
Most of the models introduced so far adopt a hard clustering approach, that is, each node belongs to one group. However, for real-world networks, it is not unreasonable to allow a node to belong to multiple groups. In this section, we will look at models that do so, by incoporating a \textit{soft clustering} approach in an SBM. Care has to be taken regarding how nodes that can belong to more than one group interact to form edges.

\subsection{Mixed membership SBM}
In the mixed membership stochastic block model (MMSBM) by \cite{abfx08}, for each node $p$, the latent variable $\Z[p]$, which contains exactly one 1, is replaced by a membership vector, also of length $K$, denoted by $\t[p]$. The elements of $\t[p]$, which represent weights or probabilities in the groups, have to be non-negative and sum to 1. Using the example in Figure \ref{fig.example}, $\Z[1]=(1~0~0)^T$ could be replaced by $\t[1]=(0.7~0.2~0.1)^T$, which roughly means that, on average, node 1 spends 70\%, 20\% and 10\% of the time in groups 1, 2 and 3, respectively. Next, each node can belong to different groups when interacting with different nodes. In order to do so, still assuming that $\mathcal{G}$ is undirected, for each dyad $(p,q)$, a latent variable $\Z[pq]$ is drawn from the multinomial distribution with probabilities $\t[p]$. As a $K$-vector containing exactly one 1, $\Z[pq]$ now represents the group node $p$ is in \textit{when interacting with $q$}. (Also drawn is $\Z[qp]$ from $\t[q]$ to represent the group node $q$ is in when interacting with $p$.) Going back to the example, if $\Z[12]=(0~0~1)^T$ and $\Z[13]=(1~0~0)^T$, which are drawn independently from $\t[1]$, node 1 belongs to groups 3 and 1, respectively, when $\Y[12]$ and $\Y[13]$ are concerned. As each $\Z[pq]$ is a $K$-vector, the collection of latent variables $\Z$ is now an $n\times{}n\times{}K$ array. The \textit{apriori} density of $\Z$ now becomes 
\begin{align}
  \pi\left(\Z|\T\right)=\Prod[n]{p\neq{}q}\left(\Z[pq]^T\t[p]\times\Z[qp]^T\t[q]\right),\nonumber
\end{align}
where $\T:=\left(\t[1]~\t[2]~\cdots~\t[n]\right)^T$ is the $n\times{}K$ matrix of membership probabilities such that $\T[pi]$ is the $i$-th element of $\t[p]$. Comparing with \eqref{eqn.graph_lik_hard}, the likelihood is
\begin{align}
  \pi\left(\Y|\Z,\C\right)=\Prod[n]{p\leq{}q}\left[\left(\Z[pq]^T\C\Z[qp]\right)^{\Y[pq]}\left(1-\Z[pq]^T\C\Z[qp]\right)^{\left(1-\Y[pq]\right)}\right].\label{eqn.graph_lik_soft}
\end{align}
We can carry out (Bayesian) inference once we specify the prior distributions. However, we shall defer this to Section \ref{sect.graph_inf}. For the derivations of the model for directed graphs, please see \cite{abfx08}. What should be noted here is that the main goal of inference is not for the pairwise latent variables $\Z$, but the mixed memberships $\T$.

Several articles built upon the MMSBM introduced. \cite{law16} proposed a scalable algorithm (Section \ref{sect.graph_inf}, \cite{fsx09} and \cite{xfs10} incorporated longitudinal modelling (Section \ref{sect.graph_longitudinal}), and \cite{kgbs13} and \cite{fcx15} focused on modelling $K$, the number of groups (Section \ref{sect.graph_K_modelled}). \cite{fxc16} observed that the assumption that $\Z[pq]$ and $\Z[qp]$ are independent is not quite realistic in real-world networks, as nodes may have higher correlated interactions towards the ones within the same groups. Therefore, they proposed a copula MMSBM for modelling these intra-group correlations.

\cite{ggms16} modified the MMSBM for recommender systems, in which the observed data is the \textit{ratings} some users give to some items (such as books or movies), and the goal of modelling and inference is to predicting user preferences. The ratings are therefore treated as the observed edges in a bipartite graph, but it is not the \textit{existence} of the these edges that is being modelled. Rather, it is the value of the ratings, that is, the edge weight, that depends on the depends on the respective mixed memberships of the users and items. By inferring these memberships as well as the block matrix, predictions on user preferences can be made \textit{for unobserved combinations of users and items}.

The MMSBM is related to, or has been compared with other models for graphs. For example, the latent feature model \citep{mgj09} deviated from the hard clustering SBM in a different way than MMSBM did, and therefore made comparison with the latter in terms of performance. For the class of latent feature models, and their practical difference with MMSBM, please see Section \ref{sect.feature}. A close connection with the latent space models \citep{hrh02,hrt07} have been drawn by \cite{abfx08}; please see Section \ref{sect.latent_space}. 

\subsection{Overlapping SBM} \label{sect.overlap}
\replaced{When applying MMSBMs, while some nodes might have genuine mixed memberships, some other nodes might have single memberships, that is, $\t[p]$ being a vector with all 0's but one 1. Such phenomenon is being addressed in another group of models, called \textit{overlapping} SBMs, in a more direct way. For example, if there are $K=2$ groups in the network, there will be nodes belonging to group 1 only, some others belonging to group 2 only, and the rest belonging to both groups simultaneously. When $K>2$, overlapping between more than two groups is allowed. Unlike the hard clustering SBMs, the groups are not disjoint anymore in these overlapping models, which therefore are an alternative to the MMSBMs, as far as soft clustering is concerned. Theoretically, there are $2^K-1$ choices of membership combinations for each node, as, for each node, there is a binary choice for belonging to each of the $K$ groups, with the only constraint that it has to have at least one group membership.}{}

\replaced{The difference with the MMSBM is noted by \cite{peixoto15b}, who used the MDL approach \citep{peixoto17a} (Section \ref{sect.micro}) for overlapping models. They first considered a variation of the Bernoulli (or Poisson) SBM, in which the memberships are relaxed in the way described in the paragraph above. They then observed that, for sparse graphs, such an overlapping model can be approximated by a non-overlapping model for an augmented graph. Each distinct membership of a single node in the original graph can be considered as a different node with a single membership in the augmented graph. Modelling and inference (Section \ref{sect.graph_inf}) are then straightforward. The expected degree for a node $p$ (in the original graph) with membership in multiple groups will be larger than that for nodes in either group, as $p$ received edges associated with each of the groups independently. Contrastly, in MMSBM, such quantity will be the weighted average of the corresponding quantities of the groups $p$ belongs to. A degree-corrected version which incoporates the DC-SBM \citep{kn11} is also derived, in which the soft clustering is achieved through hard clustering of the half-edges, rather than the hard clustering of the augmented graph in the non-degree-corrected version.}{}

\replaced{The increased number of membership choices (from $K$ to $2^K-1$) naturally brings about the increased complexity of the overlapping SBM. However, such complexity is usually not favoured in applications to real-world networks. By comparing the MDL of the non-overlapping and overlapping SBMs, \cite{peixoto15b} managed to carry out model selection, and found that the latter is more likely to overfit, and is selected as the better model only in a few cases. This finding is echoed by \cite{xks13} in the context of \textit{community detection algorithms} (Section \ref{sect.com_det}). They observed through a comparative study that, in real-world networks, each overlapping node typically belongs to 2 or 3 groups. Furthermore, the proportion of overlapping nodes is relatively small for real-world networks, usually less than 30\%.}{}

\replaced{\cite{rvw17} proposed an overlapping model that is similar to but not an overlapping SBM, because the connections between the nodes, which they called actors, are unknown in the data. Instead, available in the data are whether the actors have attended certain events. Equivalently, the data can be viewed as a bipartite network between the actors and the events. In the proposed model, clustering is applied to the actor nodes, which can belong to one or more groups, hence the overlapping nature of the model. Subsequently, memberships were inferred without the direct knowledge of edges \textit{between} the actor nodes.}{}

\section{Related methods for graphs} \label{sect.related}
In this section, two classes of models for graphs and one class of models for \textit{hypergraphs}, are reviewed, with a focus on how they work with network or graph data in different ways to SBMs. Community detection algorithms will also be mentioned, which are a class of \textit{methods} with a similar (but not identical) goal to SBMs.

\subsection{Latent feature models} \label{sect.feature}
A class of models closely related to SBMs is the latent feature models \citep{mgj09,msh11}, in which there are no longer $K$ groups but $K$ \textit{features}. For example, if Figure \ref{fig.example} represents a social network where nodes and edges correspond to people and personal connections, respectively, then the $K=3$ features could be gender (0 for female and 1 for male), whether they wear glasses, and whether they are left-handed (0) or right-handed (1). Each element of $\Z[p]$ is a binary latent variable without constraint, representing the absence or presence of a latent \textit{feature}, meaning that the sample space of $\Z[p]$ is the $2^{K}$ combinations of 0's and 1's\replaced{ (note the similarity with the number of combinations in an overlapping SBM in Section \ref{sect.overlap})}{}. Continuing with the example, if node 1 is a female who wears glasses and is right-handed, $\Z[1]=(0~1~1)^T$.

The element $\C[ij]~(1\leq{}i,j\leq{}K)$ in the matrix $\C$ represents the probability of an edge from a node with feature $i$ to a node with feature $j$. In their infinite multiple relational model (IMRM), \cite{msh11} assumed that the feature combinations are independent, which means that the probability of an edge for the dyad $(p,q)$ is
\begin{align}
  &\Pr\left(\Y[pq]=1\right|\Z,\C)=1-\Pr\left(\Y[pq]=0\right|\Z,\C)\nonumber\\
  &\qquad=1-\Prod[K]{i,j}\Pr\left(\text{No edge from $p$ with feature $i$ to $q$ with feature $j$}|\Z,\C\right)\nonumber\\
  &\qquad=1-\Prod[K]{i,j}\left(1-\C[ij]\right)^{\Z[pi]\Z[qj]}\nonumber\\
  &\qquad=1-\exp\left(\sum_{i,j}^{K}\Z[pi]\log(1-\C[ij])\Z[qj]\right)=1-\exp\left(\Z[p]^T\mathbf{P}\Z[q]\right),\label{eqn.graph_ypq_msh11}
\end{align}
where $\mathbf{P}$ is a matrix such that $\mathbf{P}_{ij}=\log(1-\C[ij])$. \cite{mgj09} specified their latent feature relational model (LFRM) in a slightly different way, by using a weight matrix $\W$ in place of $\C$ such that $\Z[p]^T\W\Z[q]$ can take any real value, and a function $\sigma(\cdot)$ that maps $(-\infty,\infty)$ to $(0,1)$ such that
\begin{align}
  \Pr(\Y[pq]=1|\Z,\W)=\sigma\left(\Z[p]^T\W\Z[q]\right).\label{eqn.graph_ypq_mgj09}
\end{align}
Not only do \eqref{eqn.graph_ypq_msh11} and \eqref{eqn.graph_ypq_mgj09} look similar to the (conditional) edge probability $\Pr(\Y[pq]=1|\Z,\C)=\Z[p]^T\C\Z[q]$ in the aforementioned version of SBM, the latent feature models can also be reduced to the SBM when only one feature is allowed, by imposing the constraint $\Z[p]^T\boldsymbol{1}_K=1$, where $\boldsymbol{1}_K$ is a $K$-vector of $1$'s.

The latent feature models should not be confused with the mixed membership models (Section \ref{sect.graph_cluster}), where a node can belong to multiple groups with weights. The practical difference is, while the SBM so far and the latent feature model allow one and multiple 1's in each $\Z[p]$, respectively, the MMSBMs allow non-binary and non-negative weights in $\t[p]$, subject to the constraint that these weights sum to 1 for each $p$.

\cite{zhou15} proposed a similar model, called the edge partition model (EPM), in which each element of $\Z$ and $\W$ is assumed to come from the Gamma distribution, resulting in a non-negative value for $\Z[p]^T\W\Z[q]$, which is assumed to be the mean rate of \textit{interaction}, for dyad $(p,q)$. Assuming that the number of interactions is Poisson distributed and that $p$ is connected to $q$ if they have interacted once, we have
\begin{align}
  \Pr\left(\Y[pq]=1|\Z,\W\right)=1-\exp\left(-\Z[p]^T\W\Z[q]\right).\label{eqn.graph_ypq_zhou15}
\end{align}
\cite{pkg12} extended the LFRM by \cite{mgj09} by introducing subclusters for the latent features in their infinite latent attribute (ILA) model. Universally, there are still $K$ features and an $n\times{}K$ binary matrix $\Z$ representing the presence or absence of latent features for the nodes. Additionally, for feature $m$, there are $K^{(m)}$ subclusters, a $K^{(m)}\times{}K^{(m)}$ weight matrix $W^{(m)}$, and an $n$-vector denoted by $\V^{(m)}$ such that $\V[p]^{(m)}$ represents the subcluster that node $p$ belongs to if it has feature $m$. If we denote the collections of subcluster vectors and weight matrices by $\V$ and $\W$, respectively, we have
\begin{align}
  \Pr\left(\Y[pq]=1|\Z,\V,\W\right)=\sigma\left(\sum_{m=1}^{K}\Z[pm]\W[{\V[p]^{(m)}\V[q]^{(m)}}]^{(m)}\Z[qm]\right),\label{eqn.graph_ypq_pkg12}
\end{align}
where $\sigma(\cdot)$ is the same as in \cite{mgj09} that maps $(-\infty,\infty)$ to $(0,1)$, such as the sigmoid function $\sigma(x)=\left(1+\exp(-x)\right)^{-1}$ or the probit function $\sigma(x)=\Phi(x)$.

Comparing the models introduced in this section, the EPM \citep{zhou15} is found to outperform the ILA model \citep{pkg12}, which in turn outperforms the LFRM \citep{mgj09}, which in turn outperforms the MMSBM \citep{abfx08} introduced in Section \ref{sect.graph_cluster}. However, \citep{msh11} did not compare their IMRM \citep{msh11} with the models here, not was there a single comparison between all these latent feature models.

%\replaced{The ILA \citep{pkg12} is found to outperfom the LFRM \citep{mgj09} when applied to NIPS coauthorship network \citep{gcpt07} and Gene interaction network \citep{jcdoqswswws09}.}{}

\subsection{Hypergraph models} \label{sect.hypergraph}
Coauthorship or collaboration networks are a popular kind of data that statistical network methods have been applied to \citep{newman01a,newman01c,newman04a,ng04,jj16}. However, the graphs are usually constructed by assigning an edge, possibly valued, to two authors if they have coauthored one or more articles. Such representation, however, does not preserve all the information \citep{nm18} and may not be very realistic. For example, pairwise edges between nodes (authors) 1, 2 and 3 could mean that each pair have collaborated separately, or that all three of them have written one or more articles as a whole, or a combination of both. Furthermore, when an article is written by, say, more than 20 authors, it is unrealistic to assume that each pair of authors know each other with equal strengths.
  
A more natural representation of such data is through the use of hypergraph. Specifically, a \textit{hyperedge} is an unordered subset of the node set $\mathcal{N}$, and when all hyperedges are node pairs, the hypergraph is reduced to a graph. In the example with the three authors, each pair having collaborated separately corresponds to 3 hyperedges: $\{1,2\}$, $\{2,3\}$ and $\{3,1\}$, whereas all three of them collaborating together corresponds to 1 hyperedge: $\{1,2,3\}$.

Hypergraph data can also be modelled with the same goal of clustering the nodes. However, it is not quite direct to extending from SBMs to ``connect a random number of two or more nodes'', making it more difficult to work with hyperedges. \cite{nm18} resorted to and extended the latent class analysis (LCA), in which the \textit{hyperedges} are clustered into the latent groups, and the memberships of the nodes can be seen as a mixture of the memberships of the hyperedges they are in.

\cite{lmwa17} considered a geometric representation of the nodes in an Euclidean space to construct a hypergraph. For ease of explanation, we assume the nodes lie on a 2-dimensional plane, and, for each node, a circle of the same radius is drawn. Then for each set of nodes that have their circles overlapped, a hyperedge is assigned. Essentially, instead of clustering the nodes into groups, this model \textit{projects} them onto an Euclidean space and infers their latent \textit{positions}.

\subsection{Latent space models} \label{sect.latent_space}
Projecting the nodes of a \textit{graph} $\mathcal{G}$ to an Euclidean space and discovering their latent positions has also been explored in the literature. \cite{hrh02} proposed the latent space model, in which the latent variable associated with node $p$, still denoted by $\Z[p]$ here, does not correspond to the group membership, but a position represented by, for example, the vector of coordinates in the Euclidean space. Then the probability of nodes $p$ and $q$ having an edge in $\mathcal{G}$, assuming it is undirected, depends on the distance between $\Z[p]$ and $\Z[q]$:
\begin{align}
  \Pr(\Y[pq]=1|\Z)=\frac{\exp\left[-d(\Z[p],\Z[q])\right]}{1+\exp\left[-d(\Z[p],\Z[q])\right]},\nonumber
\end{align}
where $d(\cdot,\cdot)$ is a distance measure, possibly with some parameters, satisfying the triangular inequality. So, probability of an edge between $p$ and $q$ decreases with the distance between $\Z[p]$ and $\Z[q]$. If covariates $\x_{pq}$ about the dyad $(p,q)$ are available, they can also be incorporated into the model:
\begin{align}
  \Pr(\Y[pq]=1|\Z,\alpha,\boldsymbol{\beta})=\frac{\exp\left[\alpha+\boldsymbol{\beta}^T\x_{pq}-d(\Z[p],\Z[q])\right]}{1+\exp\left[\alpha+\boldsymbol{\beta}^T\x_{pq}-d(\Z[p],\Z[q])\right]},\nonumber
\end{align}
where $\alpha$ and $\boldsymbol{\beta}$ are extra parameters. They noted that this model formulation is useful for handling undirected graphs because of the symmetry between $p$ and $q$. For directed graphs, they proposed
\begin{align}
  \Pr(\Y[pq]=1|\Z,\alpha,\boldsymbol{\beta})=\frac{\exp\left[\alpha+\beta^T\x_{pq}+\Z[p]^T\Z[q]/|\Z[q]|\right]}{1+\exp\left[\alpha+\beta^T\x_{pq}+\Z[p]^T\Z[q]/|\Z[q]|\right]},\nonumber
\end{align}
where the asymmetric term $\Z[p]^T\Z[q]/|\Z[q]|$ is the signed magnitude of $\Z[p]$ in the direction of $\Z[q]$.

real-world networks usually exhibit transitivity, which means that, if both nodes A and B are connected to node C, then A and B are likely to be connected. Another common phenonmenon is homophily, which means that nodes with similar attributes are more likely to be connected. They are accounted for by the above model through the use of latent space and dyad-specific covariates $\x_{pq}$, respectively. \cite{hrt07} proposed an extension in the form of a latent space \textit{cluster} model, by considering the \textit{apriori} distribution of the latent positions. Specifically, for each node $p$, $\Z[p]$ is assumed to be drawn from a mixture of $K$ Gaussian distributions, each of which has a different mean and covariance matrix to represent a different group/cluster. In this way, the clustering of the nodes are accounted for explicitly.

\cite{abfx08} noted the similarity between their MMSBM and the latent space models. For generating an edge between nodes $p$ and $q$, the terms $\Z[pq]^T\C\Z[qp]$ and $\Z[p]^T\mathbf{I}\Z[q]$ are involved in the former and the latter, respectively, where $\mathbf{I}$ is an identity matrix. They have also compared their performances when being applied to the same set of data.

A recent development with latent space models is by \cite{sh19}, who first used spectral clustering \citep{vonluxburg07} to project, or embed, the graph to a $d$-dimensional Euclidean space. Then, a Gaussian mixture model, with $K$ components, is fit to these spectral embeddings. Their novelty is the estimation of $d$ \textit{and} $K$ simultaneously (Section \ref{sect.graph_K_modelled}) in their inference algorithm. For more details on spectral clustering and its relation to SBM, please see \cite{rcy11}.

\subsection{Community detection} \label{sect.com_det}
Without the pretext of statistical or probabilistic modelling, community detection can be the goal of analysing a network, which is to cluster nodes so that the edge density is high within a group and low between groups. This concept is also called assortativeness. In the context of SBMs, this means $\C[ii]~(i=1,2,\ldots,K)$ is high while $\C[ij]$ is low for $j\neq{}i$. As mentioned in Section \ref{sect.sto_equ}, this is not guaranteed by the concept of stochastic equivalence alone. While SBMs can find communities with high within-group edge densities, they are in fact a more general method that allow other types of structure in the network to be found \citep{gs09,mmfh13}.

The above effect is illustrated by, for example, the difference between the DC-SBM by \cite{kn11} and the original version, when applied to a real-world network with $K=2$. While the former accounted for the variation in the degree and managed to discover the two communities, the latter put the highly connected nodes together in one group, the rest in another. In bigger networks with nodes on the \textit{periphery} of the network, that is, they are only connected to one or a few nodes which are more central to the network, these peripheral nodes will be put together in a ``miscellaneous'' group with a low edge density, under the original SBM, instead of the same groups as the more central nodes they are connected to.

\subsubsection{Assortative SBM} \label{sect.assort}
One way of achieving community detection is to modify the SBMs to align with this goal. In the assortative (or affinity) SBM \citep{gmgfb12,law16}, a constraint is imposed that $\C[ij]=\delta$ for $i\neq{}j$, where $\delta$ is a parameter presumed to be smaller than $\C[ii]$. While reducing the number of parameters in $\C$ from $K(K+1)/2$ (in the case of undirected graphs, $K^2$ in the case of directed ones) to $K+1$ may not significantly reduce the computational cost unless $K$ is large, it implies assortativeness. However, it should be noted that incorporating assortativeness in SBMs is not a universal solution. For example, it is not sensible when bipartite networks are modelled, in which connectivity is high between groups but zero within groups. Therefore, caution should be taken whenever an assortative SBM is used, although the stochastic gradient method by \cite{law16} should be easily generalisable to a non-assortative model.

\replaced{\cite{ls19b} proposed a regularised SBM which extends the DC-SBM to control the desired level of assortativeness. The expected number of edges for the dyad $(p,q)$ is now $\phi_p\phi_q\Z[p]^T\C\Z[q]$ if $\Z[p]=\Z[q]$, $(k_p-\phi_p)(k_q-\phi_q)\Z[p]^T\C\Z[q]$ otherwise, where $k_p$ is the degree of node $p$. While a different expected number of edges is allowed according to the group memberships, the parameter $\phi_p$ is regulated by a parameter $h$, which is a number between 0 and 1, according to $\phi_p=\max(hk_p,1)$. The tuning parameter $h$ is not estimated but varied, to give differnt clustering results corresonding to different levels of assortativeness. A high value of $h$ leads to a more assortative partition and, in the application, recovers the same known factions in the karate club network as in \cite{kn11}.}{}

\replaced{The assortative models introduced so far are actual SBMs and not merely related methods for graphs. It should be noted that they are introduced here because of the proximity to the goal of community detection.}{}

\subsubsection{Non-probabilistic and modularity methods}
Another way of achieving community detection is to step away from SBMs, and apply methods \replaced{which are }{}not based on statistical or probabilistic \textit{modelling} but mainly on heuristics, and are usually iterative in nature. For example, in the label propogation algorithm by \cite{rak07}, initially each node is randomly assigned to one of the $K$ groups. Then each node takes turn to join the group to which the maximum number of its neighbours belong (with ties broken uniformly randomly). The iterative process continues until no node changes its group membership anymore. Other methods, for example, are based on the edge betweenness centrality measure \citep{gn02}, random walks on the graph \citep{pl06}, and network flows and information-theoretic principles \cite{rab09}.

One very popular framework of community detection is the use, and optimisation, of the modularity of a network, by the highly-cited \citep{ng04}. Assuming an undirected (but possibly valued) graph, the formulae of the modularity, denoted by $Q$, is
\begin{align}
  Q=\frac{1}{2M}\sum_{p,q}^n\left[\Y[pq]-\frac{k_pk_q}{2M}\right]\one{\Z[p]=\Z[q]},\label{eqn.modularity}
\end{align}
where $k_p=\sum_{q=1}^n\Y[pq]$ is the \textit{degree} of node $p$. As it can be interpreted as the number of edges within groups minus the expected number of such edges, the modularity is a very useful measure for how ``good'' the clustering is, and numerous algorithms have been proposed for its optimisation \citep{cnm04,newman06a,newman06b,wt07,bgll08}. Moreover, it can be computed for results not based on modularity optimisation, and compared between all methods. \cite{cnm04} suggested that, in practice, a modularity above 0.3 is a good indicator of significant community structure.

\cite{newman16} established a connection between modularity optimisation and the DC-SBM (Section \ref{sect.poisson}).\replaced{ Specifically, maximising the likelihood of a special case of the DC-SBM is equivalent to maximising a generalised version of the modularity. This is also noted by, for example, \cite{ls19a}, when they tackled the resolution limit problem, which is reviewed below.}{}

\subsubsection{Resolution limit}
Community detection algorithms are in general easy to implement, and likely to be faster than applying an SBM. However, its disadvantages includes that different initial configurations may lead to different results even under the same algorithm, and that different results give vastly different results. More importantly, modularity optimisation methods suffer from a resolution limit \citep{fb07,gdc10,lf11}. This means that by maximising the modularity, smaller groups or clusters cannot be detected, especially for large graphs. For example, performing community detection on a network of 1 million nodes may result in the smallest group having 500 nodes, but the algorithm performed on the whole network cannot go deeper to further cluster these 500 nodes (or that modularity will decrease by doing so). \replaced{The need of a possible additional round of community detection is not ideal as it should have been a systematic part of the initial community detection.}{}

\replaced{The above issue of resolution limit is tackled by recent works in various ways. \cite{cks14} proposed an alternative to modularity, called modularity density, which theoretically resolves the resolution limit problem, and empricially improves results significantly when being applied to community detection algorithms. \cite{ls19a} proposed an agglomerative community detection algorithm to detect communities at multiple scales, thus avoiding the resolution limit issue. }{}\cite{peixoto13} noted the connection with the resolution limit for SBMs\replaced{, and went on to resolve the issue in the hierarchical model \citep{peixoto14b} (Section \ref{sect.micro}).}{}

\subsubsection{Miscellaneous}
Community detection is, in itself, a largely studied topic in the literature\replaced{, and a few important reviews should be referred to for further exploration of the topic}{}. \cite{fortunato10} provided an introduction and comprehensive review of community detection, while \cite{abbe18} surveyed the developments that establish the fundamental limits for community detection in SBMs.\replaced{ \cite{sdrl17} reviewed a spectrum of algorithms according to their different motivations that underpin community detection, with the aim of providing guildlines for selecting appropriate algorithms for the given purposes. In addition to providing an overview of algorithms based on modularity optimisation, \cite{cpsl19} reviewed recent advances on two specific topics, namely community detection for time evolving networks, and immunisation strategies in networks with overlapping and non-overlapping community structure. The former is, in the context of SBMs, reviewed in Section \ref{sect.graph_longitudinal}, while the latter is useful for, for example, targeting a small group of nodes to prevent the spread of epidemics in networks.}{}

\section{Inference approach}\label{sect.graph_inf}
In this section, the general framework of inference for SBMs is reviewed, in which there are two main classes of methods: Monte Carlo (Section \ref{sect.graph_inf_monte_carlo}) and variational (Section \ref{sect.graph_inf_variational}), to answer question \ref{q.inference} in Section \ref{sect.intro}. Greedy methods will be mentioned in Section \ref{sect.graph_inf_misc}, while methods for predicting or correcting the edges or non-edges will be presented in Section \ref{sect.graph_inf_pred}. While the general approach is discussed here, the more specific models or algorithms for estimating or selecting the number of groups $K$ will be mentioned in Section \ref{sect.graph_K}. Nevertheless, inference and the issue of $K$ are, most of the time, intertwined with each other, and are presented in separate sections here for clarity.

We first observe that, by combining \eqref{eqn.graph_lik_hard} and \eqref{eqn.graph_z_hard}, inference is possible by the frequentist approach. This can be achieved via direct maximisation of likelihood or the expectation-maximisation (EM) algorithm \cite{dlr77}, both of which are illustrated in \cite{sn97} for a simple case where $K=2$. However, we will focus on the more popular and arguably more powerful Bayesian approach here, as did \cite{ns01}. What remains is assigning priors to $\C$ and $\t$ before inference can be carried out. We assume each element of $\C$ has an independent Beta prior, that is $\C[ij]\sim\text{Beta}\left(\A[ij],\B[ij]\right)$, where $\A$ and $\B$ are $K\times{}K$ matrices with all positive hyperparameters. We also assume $\t$ arises from the Dirichlet$\left(\a\boldsymbol{1}_K\right)$ distribution, of which the parameter $\a$ comes from a Gamma$(a,b)$ prior.

The joint posterior of $\Z$, $\t$, $\C$ and $\a$, up to a proportionality constant, is
\begin{align}
  &\pi\left(\Z,\t,\C,\a|\Y\right)\propto\pi\left(\Y,\Z,\t,\C,\a\right)\nonumber\\
  &\quad=\pi\left(\Y|\Z,\t,\C,\a\right)\times\pi\left(\Z|\t,\C,\a\right)\times\pi\left(\t|\C,\a\right)\times\pi\left(\C,\a\right)\nonumber\\
  &\quad=\pi\left(\Y|\Z,\C\right)\times\pi\left(\Z|\t\right)\times\pi\left(\t|\a\right)\times\pi\left(\C\right)\times\pi\left(\a\right)\nonumber\\
  \begin{split}
  &\quad\propto\Prod[n]{p<q}\left[\left(\Z[p]^T\C\Z[q]\right)^{\Y[pq]}\left(1-\Z[p]^T\C\Z[q]\right)^{\left(1-\Y[pq]\right)}\right]\times\Prod[n]{p=1}\Z[p]^T\t\\
  &\qquad\times\left[\Gamma\left(K\a\right)\one{\sum_{i=1}^K\theta_i=1}\Prod[K]{i=1}\frac{\theta_i^{\a-1}}{\Gamma\left(\a\right)}\right]\\
  &\qquad\times\Prod[K]{i\leq{}j}\left[\C[ij]^{\A[ij]-1}\left(1-\C[ij]\right)^{\B[ij]-1}\right]\times\a^{a-1}e^{-b\a}.
  \end{split}\label{eqn.graph_inf_joint_hard}
\end{align}

Under the soft clustering approach in Section \ref{sect.graph_cluster}, the weight vectors $\{\t[1],\t[2],\ldots,\t[K]\}$ are assumed to be i.i.d. according to the Dirichlet$(\a\boldsymbol{1}_K)$ distribution. This yields the joint posterior of $\Z$, $\T$, $\C$ and $\a$:
\begin{align}
  &\pi\left(\Z,\T,\C,\a|\Y\right)\propto\pi\left(\Y|\Z,\C\right)\times\pi\left(\Z|\T\right)\times\pi\left(\T|\a\right)\times\pi\left(\C\right)\times\pi\left(\a\right)\nonumber\\
  \begin{split}
    &\quad\propto\Prod[n]{p<q}\left[\left(\Z[pq]^T\C\Z[qp]\right)^{\Y[pq]}\left(1-\Z[pq]^T\C\Z[qp]\right)^{\left(1-\Y[pq]\right)}\right]\\
    &\qquad\times\Prod[n]{p<q}\left(\Z[pq]^T\t[p]\Z[qp]^T\t[q]\right)\times\Prod[n]{p=1}\left[\Gamma\left(K\a\right)\one{\t[p]^T\boldsymbol{1}_K=1}\Prod[K]{i=1}\frac{\T[pi]^{\a-1}}{\Gamma\left(\a\right)}\right]\\
    &\qquad\times\Prod[K]{i\leq{}j}\left[\C[ij]^{\A[ij]-1}\left(1-\C[ij]\right)^{\B[ij]-1}\right]\times\a^{a-1}e^{-b\a}.
  \end{split}\label{eqn.graph_inf_joint_soft}
\end{align}
Comparing this with \eqref{eqn.graph_inf_joint_hard} illustrates why the hard clustering approach is preferred in the literature of SBMs. As $K$ is a lot smaller than $n$ usually (hence the purpose of clustering), the computational cost mainly depends on the number of latent variables $\Z$. In the hard and soft clustering approaches, this amounts to $O(n)$ and $O(n^2)$ iterations, respectively. The quadratic computation cost means that a simple Gibbs sampler is not very scalable in soft clustering \citep{msh11}.

\subsection{Monte Carlo methods} \label{sect.graph_inf_monte_carlo}
If algorithmic simplicity is preferred to computational efficiency, inference can be carried out in a straightforward way via Markov chain Monte Carlo (MCMC). More specifically, a simple regular Gibbs sampler can be used, where all the parameters and latent variables (except $\a$) can be updated via individual Gibbs steps; see, for example, \cite{ns01} and \cite{lw18}. In quite a few cases, a Gibbs sampler is natural when $K$ is being modelled (Section \ref{sect.graph_K_modelled}), and this is used by both SBMs \citep{ty11,ty14,pkg12,fcx15,zhou15} and a latent feature model \citep{mgj09}. Other articles on SBMs that use MCMC include \cite{yczgj11,dbs13,mmfh13,law16,len17}, and \cite{peixoto17a,peixoto17b,peixoto18a}. \cite{msh11} proposed using Hamiltonian Monte Carlo (HMC) for (a transformation of) each element in $\C$, in which the gradient of the log-posterior is utilised. In their dynamic SBM, \cite{xh13} incorporated a sequential Monte Carlo (SMC) step with a label-switching algorithm by \cite{kn11} (explained below).

If computational efficiency is the focus of the MCMC algorithm, careful considersations are required so as not to waste computational time on naive moves. \cite{msh11} noted that, in a latent feature model, the possible combinations of latent features is $2^{Kn}$, compared to the $K^n$ possible combinations of group memberships in an SBM, and so standard Gibbs samplers are unlikely to be scalable. Therefore, in their algorithm, certain types of moves are employed so that the computational complexity increases linearly with the number of \textit{edges}, not of \textit{node combinations}. \cite{law16} made a similar claim for their stochastic gradient MCMC algorithm, in which only a small mini-batch of the nodes is required in each iteration, to greatly reduce the computational overhead. These algorithms benefit from the fact that large graphs are usually sparse in reality.

\subsubsection{Collapsed SBMs and efficient moves} \label{sect.graph_inf_collapsed}
While both \cite{mmfh13} and \cite{peixoto17a,peixoto17b} integrated $\C$ out in their respective SBMs, they proposed different efficient moves in their MCMC algorithms. In \cite{mmfh13}, one of the following four moves is selected randomly uniformly and performed in each iteration. The first one is a Gibbs move for a randomly selected node. The second is a move that selects two groups at random and proposes to reassign all the nodes in these two groups, in a scheme described by \cite{nf07}. The last two are moves that affect the number of groups alongside the memberships, as $K$ is treated as a parameters and estimated. \cite{mmfh13} also proposed a method to deal with the issue of label-switching, which is circumvented in the spectral clustering model by \cite{sh19}.

\cite{peixoto17a,peixoto17b} applied a single-node move proposed by \cite{peixoto14a} that works as follows. For node $p$, whose membership $\Z[p]$ is to be updated, a neighbour $q$ is selected randomly uniformly, whose membership is, say, $\Z[q]=i$. Then node $p$ is proposed to move to a group $j$ (which could be the same as $i$) with probability proportional to $\E[ij]$, that is, the number of edges between groups $i$ and $j$. Note that this is different from proposing to move $p$ to a group $j$ with probability proportional to the number of neighbours of $p$ in group $j$. Another move is also proposed for merging groups, as $K$ is being modelled (Section \ref{sect.graph_K_modelled}).

\subsection{Variational methods} \label{sect.graph_inf_variational}
An alternative to the Monte Carlo methods is the class of variational expectation-maximisation (VEM) methods. The principle of these algorithms is to first provide a lower bound to the marginal log-likelihood, in which the latent variables ($\Z$ here) are integrated out, by the use of an approximate \textit{variational} distribution. This lower bound is then tightened or maximised, with respect to the model parameters ($\t$, $\C$ and $\a$ here) as well as those of the variational distribution. We illustrate this using the Bernoulli SBM. By writing $\e=\{\t,\C,\a\}$, for any distribution $Q$ of the latent variables $\Z$, we have
\begin{align}
  \log\pi(\Y|\e)&=\log\int\frac{\pi(\Y,\Z|\e)}{Q(\Z)}Q(\Z)d\Z\nonumber\\
  &\geq\int\log\frac{\pi(\Y,\Z|\e)}{Q(\Z)}Q(\Z)d\Z\nonumber\\
  &=\mathbb{E}_Q\left[\log\pi(\Y,\Z|\e)-\log{}Q(\Z)\right].\label{eqn.graph_inf_lower_bound}
\end{align}
The second line is due to Jensen's inequality as the logarithm function is concave. As $\log\pi(\Y|\e)=\int\log\pi(\Y|\e)Q(\Z)d\Z$, the difference in the inequality is
\begin{align}
  &\log\pi(\Y|\e)-\int\log\frac{\pi(\Y,\Z|\e)}{Q(\Z)}Q(\Z)d\Z\nonumber\\
  &\qquad=\int\left[\log\pi(\Y|\e)-\log\frac{\pi(\Y,\Z|\e)}{Q(\Z)}\right]Q(\Z)d\Z\nonumber\\
  &\qquad=\mathbb{E}_Q\left[\log\frac{Q(\Z)}{\pi(\Z|\Y,\e)}\right],\nonumber
\end{align}
which is the Kullback-Leibler (K-L) divergence of $\pi(\Z|\Y,\e)$ from $Q(\Z)$, denoted by $D_{KL}\left(Q(\Z)~||~\pi(\Z|\Y,\e)\right)$. Therefore,
\begin{align}
  \log\pi(\Y|\e)=\mathbb{E}_Q\left[\log\pi(\Y,\Z|\e)-\log{}Q(\Z)\right]+D_{KL}\left(Q(\Z)~||~\pi(\Z|\Y,\e)\right).\label{eqn.graph_inf_vem}
\end{align}
As the marginal likelihood on the left-hand side is constant to the choice of $Q$, maximising the first term with respect to $Q$ and $\e$ is equivalent to minimising the K-L divergence, thus improving the approximation. While the ideal choice of $Q(\Z)$ is the conditional distribution $\pi(\Z|\Y,\e)$ such that the K-L divergence is $0$, the latter is usually intractable in the models discussed here, and so the best tractable choices of $Q(\Z)$ are being sought. Usual choices are such that $Q(\Z)$ is factorisable, making analytical calculations of the lower bound possible. Finally, the lower bound is iteratively maximised with respect to $\e$ in the M step, and (the parameters of) $Q$ in the E step, in an EM algorithm.

The factorisable variational distributions and the iterative steps have been illustrated by, for example, \cite{kks06} and the following references in this section. \cite{gmgfb12} and \cite{kgbs13} considered stochastic optimisation in place of EM algorithms, while \cite{mrv15} considered variants of the M step. \cite{hkk16} opted for belief propagation as the alternative approach to obtaining the variational distribution $Q$. 

That the variational approach is adopted is due to several reasons in different articles. \cite{abfx08} argued that in the MMSBMs \citep{abfx08,fsx09,xfs10} it outperforms (with computational cost $O(nK+2K)$) the corresponding MCMC methods (with computation cost $O(n^2)$). Similarly, \cite{vhs13} argued that their algorithm for SBM is more scalable $(O(n))$ compared to latent space models $(O(n^2))$ (Section \ref{sect.latent_space}). In some other cases, it facilitates the computation of a criterion, such as the ICL $\pi(\Y,\Z)$ \citep{lba12,mrv15,hkk16,mm17}, or the observed data log-likelihood $\pi(\Y|\e)$ \citep{ysjkmzzz14}. This criterion can be directly used for model selection, or equivalently selecting $K$ (Section \ref{sect.criteria}).

\subsection{Greedy methods and others} \label{sect.graph_inf_misc}
While most articles in the literature have used either Monte Carlo methods or variational methods, there are a few exceptions. In the DC-SBM \citep{kn11}, the log-likelihood can be obtained by summing the logarithm of \eqref{eqn.graph_ypq_kn11} over all possible dyads. Instead of integrating out $\C$ and $\boldsymbol{\phi}$, they adopted a frequentist approach and substituted their respective MLEs to obtain the objective function $\pi(\Y|\Z)$, which is the basis of their label-switching algorithm. In each step, each node is proposed to move from one group to another, and selected is the move that will most increase or least decrease $\pi(\Y|\Z)$. Once all nodes have been moved and the group memberships $\Z$ are according updated, the objective function is calculated for all the steps this greedy algorithm passed through, and the one with the highest objective is selected as the initial state of another run of the algorithm. The algorithm is stopped when there is no further increase in the objective function. This label-switching algorithm is also incorporated by \cite{xh13} in the inference algorithm for their dynamic SBM, which will be discussed in Section \ref{sect.graph_longitudinal}.

\cite{cl15} also proposed a greedy step in their inference algorithm. Before doing so, required is the ICL, $\log\pi(\Y,\Z)$, by integrating out the parameters $\e=\{\t,\C,\a\}$. They worked out an asymptotic version via certain approximations, as well as an exact version under certain priors. The ICL then becomes the objective function in their greedy optimisation algorithm. Similar to \cite{kn11}, in each step, each node is proposed to move from one group to another, and selected is the move that will most increase the ICL. However, if no proposed move results in an increase, $\Z[p]$ remains unchanged. The algorithm terminates again if there is no further increase in the objective. 

\cite{yan16} combined the work by \cite{kn11} and \cite{cl15}, by approximating the ICL for the DC-SBM. As the focus is selecting the number of groups by comparing the objective under different values of $K$, \cite{yan16} only provided the calculations for the ICL in the absence of an inference algorithm for the parameters and/or the latent variables.

\replaced{In the work by \cite{vv18}, they did not consider the inference algorithm and were mainly concerned with the posterior mode, that is, the value of $\Z$ that maximises the posterior density in \eqref{eqn.graph_inf_joint_hard}, or the collapsed version thereof according to \cite{mmfh13}. They found that this mode converges to the true value of $\Z$ (as $n\rightarrow\infty$), under the frequentist setup that there \textit{is} such a true value, and with the condition that the expected degree is at least of order $(\log{}n)^2$. The reason that the priors of $\Z$ (and of other parameters) are used even in a frequentist setup is that such priors actually play a part in establishing the convergenece. Furthermore, \cite{vv18} established a connection of such maximised posterior density, termed Bayesian modularity, with the likelihood modularity defined by \cite{bc09a} (Section \ref{sect.criteria}).}{}

\subsection{Missingness and errors} \label{sect.graph_inf_pred}
In some cases, the goal of applying an SBM is not (only) the inference of the group memberships $\Z$, but on dealing with partially or errorfully observed graphs. For example, there is no information regarding the interactions between some dyads. Inferring these edges or non-edges, with some (un)certainty, then becomes the goal of inference, and is called the \textit{link prediction} problem in the literature. In the context of SBMs, this has been looked into by, for example, \cite{gs09,zhou15,vmgs16,zwlz17,vpsg18,sbknm19} and \cite{tggs19}, and also by \cite{mgj09} for latent feature models. Closely related is the issue with errors in the observed graphs, where the spurious edges may lead to wrong conclusions. Here, \textit{network reconstruction} becomes another goal of inference. This has been studied by, for example, \cite{gs09,pstv15} and \cite{peixoto18r}, in the context of SBMs.

%\replaced{\textbf{\cite{hpf16}: Node prediction, not link prediction}}{}

\section{Number of groups} \label{sect.graph_K}
As the main goal of SBMs is to cluster nodes into groups or communities, without a given number of groups $K$, it is difficult to evaluate the likelihood and infer the group memberships $\Z$. Unless there is prior information on $K$, it might not be objective to carry out inference for one fixed value of $K$. Solutions usually come in two main directions. The first is to fit an SBM to multiple values of $K$, with a measure computed to quantify the goodness of fit, followed by selecting the optimal $K$. Works in this direction will be reviewed in Section \ref{sect.criteria}. The second is to model $K$ as yet another parameter and estimate it in the inference, most likely by transdimensional inference algorithms. Works in this direction will be reviewed in Section \ref{sect.graph_K_modelled}. Together, these methods can be seen as an answer to questions \ref{q.selection} and \ref{q.K} in Section \ref{sect.intro} simultaneously.

In some articles, the number of groups is fixed, rather than being estimated. \cite{sn97} set $K=2$, enabling the MLE calculations and working out the associated EM algorithm. \cite{tallberg05} and \cite{kn11} used $K=2$ and $K=3$, respectively, for their data but acknowledged the issue of assuming that $K$ is given, which is usually not the case in practice. \cite{yczgj11} focussed on the dynamic structure of social networks, and fixed $K$ to 2 for two of their datasets, and to 3 for a third dataset, by using prior knowledge on the three sets of data. Similarly, \cite{vhs13} set $K=5$ by following external relevant practices of using five groups for their data. \cite{xh13} fixed $K=7$ for their Enron email network data, also using prior knowledge on the classes the nodes (employees) belonged to.  In their model for recommender systems, \cite{ggms16} reported the results for 10 groups of users and 10 groups of items, and found no differences in performance when either number of groups is increased. The data analysed by \cite{zmn17} contained individual attributes that allowed them to divide the nodes into $K=3$ groups, and their focus was on showing that their dynamic SBM obtained results closer to the ground truth than the static counterpart did.\replaced{ When \cite{vv18} illustrated the connection of their Bayesian modularity with the likelihood modularity, through the proximity of the empirical results to those by \cite{bc09a}, they used both $K=2$ and $K=4$ in the application to the same set of data.}{}

\subsection{Criteria and model selection} \label{sect.criteria}
\cite{ns01} made separate fits to their data using $K=2,3,4,5$, and compared the \textit{information} as well as a parameter representing the ``clearness'' of the block structure, to aid their decision on $K=3$. This set the precedent to using some kind of criterion to select the optimal $K$. 

A measure previously mentioned that requires no modelling is the modularity in \eqref{eqn.modularity}, which has been compared with the profile log-likelihood by \cite{bc09a}, who called it a likelihood modularity. To obtain this criterion, first observe that, in \eqref{eqn.graph_lik_hard_re}, the MLE of $\C[ij]$ is simply $\hat{\C}_{ij}=\E[ij]/\N[ij]$, the edge density between groups $i$ and $j$. This estimate can then be plugged in \eqref{eqn.graph_lik_hard_re} to obtain
\begin{align}
  \log\pi(\Y|\Z,\hat{\C})=\sum_{i,j}^K\N[ij]\left[\frac{\E[ij]}{\N[ij]}\log\frac{\E[ij]}{\N[ij]}+\left(1-\frac{\E[ij]}{\N[ij]}\right)\log\left(1-\frac{\E[ij]}{\N[ij]}\right)\right].\nonumber
\end{align}
While \cite{bc09a} did not use the profile log-likelihood mainly for model selection, it can be seen as a precedent in moving from the non model-based modularity to a criterion based on SBM.

Another criterion used in earlier works is the Bayesian information criterion (BIC) \citep{abfx08,fsx09,xfs10}. However, according to \citep{lba12} and \cite{cl15}, it is not tractable in most cases as it depends on the \textit{observed data} log-likelihood $\log\pi(\Y|\e)$, where $\e$ represents the collection of parameters, and is known to misestimate $K$ in certain situations \citep{ysjkmzzz14}. \cite{yan16}, however, drew connection with the minimum description length (MDL) \citep{peixoto13} and dervied a more principled BIC.

Instead of evaluating the BIC, \cite{ysjkmzzz14} focussed on approximating the observed data log-likelihood $\log\pi(\Y|\e)$. They applied what they called belief propagation in their EM algorithm, in order to approximate $\log\pi(\Y|\e)$ from $\log\pi(\Y|\Z,\e)$, which is derived for both DC-SBM and its original counterpart. By doing so, a (log-)likelihood ratio can be computed, denoted by $\log\pi(\Y|\e_1)-\log\pi(\Y|\e_2)$, where $\e_1$ and $\e_2$ correspond to the parameters under the DC-SBM and the original version, respectively. With its asymptotic behaviour derived, a likelihood ratio test can be performed to answer the question of whether degree correction should be applied to the data in hand.

While $\log\pi(\Y|\e)$ or its approximation is also being used by, for example, \cite{dkmz11} (who termed it free-energy density in statistical physics terms), more works were based on the ICL $\log\pi(\Y,\Z)$, which is equivalent to the MDL \citep{peixoto14b,nr16} under compatiable assumptions. In order to obtain it, we start with the quantity $\log\pi(\Y,\Z|\e)$ in \eqref{eqn.graph_inf_vem}, and attempt to integrate $\e$ out:
\begin{align}
  \log\pi(\Y,\Z)=\log\int\pi(\Y,\Z|\e)\pi(\e)d\e,\nonumber
\end{align}
However, this is usually not tractable \citep{cl15}. \cite{dpr08} provided an approximation for directed graphs:
\begin{align}
  \log\pi(\Y,\Z)\approx\underset{\e}{\max}\log\pi(\Y,\Z|\e)-\frac{K^2}{2}\log(n(n-1))-\frac{K-1}{2}\log(n).\label{eqn.icl_approx}
\end{align}
\cite{dpr08} also provided an algorithm for the optimisation of the first term on the right hand side. This approximate ICL has been adopted by, for example, \cite{mrv15,mm17}, and \cite{sbknm19}.

The above approximation, however, can be improved by noting that some parameters in $\e$, such as $\C$ and $\t$, can be integrated out by considering conjugate priors. This is exactly the case in \eqref{eqn.graph_ypq_mmfh13}, which in fact is equivalent to $\log\pi(\Y,\Z|\e)$ in \eqref{eqn.graph_inf_vem} but with $\e=(\a,\b,\gamma)$. Noting such potential, \cite{lba12} and \cite{cl15} integrated out parameters whose dimensions depend on $K$ or $n$, and fixed the values of the remaining scalar parameters in $\e$, but parted ways in their subsequent novelty. While \cite{cl15} derived an exact ICL, \cite{lba12} proposed to approximate the marginal log-likelihood $\log\pi(\Y)$, using the version of \eqref{eqn.graph_inf_vem} without $\e$. After the convergence of the variational algorithm (Section \ref{sect.graph_inf_variational}), the first term on the right hand side of \eqref{eqn.graph_inf_vem} can be computed by plugging in the estimated $\Z$. This maximised lower bound $\mathbb{E}_Q\left[\log\pi(\Y,\Z)-\log{}Q(\Z)\right]$ is then used by \cite{lba12} as the approximation of the marginal log-likelihood, under the assumption that the K-L divergence is close to zero and does not depend on $K$. This is also being adopted by \cite{ajc15}, who calculated the Bayes factor of one model to another, according to the difference of their respective approximate marginal log-likelihoods. Similarly, \cite{hkk16} provided an approximation to the marginal log-likelihood, this time an asymptotic one, called the fully factorised information criterion (F${}^2$IC).

\cite{wb17} also investigated log-likelihood ratio similar to \cite{ysjkmzzz14}, but this time for an underfitting/overfitting $K$ against the true $K$. Using its asymptotic results, they derived a penalty term $\lambda\frac{K(K+1)}{2}n\log{}n$, where $\lambda$ is a tuning parameter, and subsequently a penalised (log-)likelihood criterion. However, \cite{hqyz19} argued that this penalty tends to underestimate $K$, and therefore proposed a lighter penalty $\lambda{}n\log{}K+\frac{K(K+1)}{2}\log{}n$ in their \textit{corrected} BIC.

Other criteria used include the mean log-likelihood of held-out test data \citep{gmgfb12,dbs13,law16}, termed perplexity by the latter, as the criterion for selecting $K$. Finally, \cite{len17} considered the ratio of the sum of the squared distances in the $k$-means clustering for all nodes in different groups, to the sum of squared distances between all node pairs, in what they called the Elbow plot, to determine the number of groups.

\cite{vpsg18} investigated an empirical approach to model selection, which is to maximise the performance in link prediction (Section \ref{sect.graph_inf_pred}). They found that the results are sometimes inconsistent with those by a criterion formulated in a similar way to those described above, which is the posterior likelihood $\pi(\Z|\Y)$ in their case.

\subsection{Modelling} \label{sect.graph_K_modelled}
The formulation of the SBM and the inference algorithm in Section \ref{sect.graph_inf} relies on $K$ being specified beforehand. To increase the flexibility of the model, $\t$ can be assumed to arise not from a Dirichlet distribution but from a (Hierarchical) Dirichlet process, which will be introduced in Section \ref{sect.topic_K_hdp}, as it was first used in topic modelling. In this way, $K$ becomes a random quantity generated by the process, and potentially an infinite number of groups is allowed. By incorporating the Dirichlet process in the SBM, as did \cite{kks06,msh11,ty11,ty14,kgbs13} and \cite{fcx15}, $K$ can be estimated along other parameters and latent variables.

A similar structure for models which are not SBMs can be incorporated, so that $K$ can be modelled and inferred. In the latent feature model by \cite{mgj09}, an Indian buffet process is used, while a hierarchical Gamma process is used in the edge partition model by \cite{zhou15}.

One major issue with modelling and inferring $K$ is that, in the SBMs with $\C$ and/or $\t$ \textit{not} integrated out, the number of parameters changes with $K$. As we have seen in \eqref{eqn.graph_ypq_mmfh13} and \eqref{eqn.graph_ypq_micro}, as well as in Section \ref{sect.criteria}, in some cases they can be marginalised, usually with conjugate priors. Having a likelihood in which the number of parameters is constant to $K$ greatly facilitates the associated inference algorithm, regardless of whether $K$ is modelled or selected by a criterion. Therefore, in recent years, these models have been preferred to the aforementioned models with a nonparametric \textit{process} for $K$. In the case where $K$ is being modelled, estimation is very often carried out using MCMC, and care has to be taken when $K$ grows or shrinks in the algorithm. Both \cite{mmfh13} and \cite{sh19} included two moves that allow so. The first move proposes to add or remove an empty group, while the second proposes to merge two groups into one \textit{or} split one into two.

\cite{peixoto14a} also allowed $K$ to be modelled, and proposed a progressive way of merging groups. To explain such way, first consider each group $i$ in the original graph, denoted by $\mathcal{G}^{(0)}$, as a \textit{node} in a graph one level above, denoted by $\mathcal{G}^{(1)}$. Similarly, the total number of edges $\E[ij]^{(0)}$ between groups $i$ and $j$ in $\mathcal{G}^{(0)}$ is now the valued edge $\Y[ij]^{(1)}$ between nodes $i$ and $j$ in $\mathcal{G}^{(1)}$. Then proposing to merge groups in $\mathcal{G}^{(0)}$ can be seen as proposing to move the membership of a node in $\mathcal{G}^{(1)}$, and this can be carried out in a similar fashion to that described in Section \ref{sect.graph_inf_collapsed}.

Similar to \cite{mmfh13}, \cite{nr16} also considered the possibility of empty groups, through modelling $K$ explicitly as a parameter. Specifically, $K$ represents the number of groups that the nodes can potentially occupy, not the number they actually do according to $\Z$. Such possibility of empty groups ensure the joint posterior of $\Z$ and $K$, with other parameters integrated out in their Poisson SBM, to be computed correctly:
\begin{align}
  \pi(\Z,K|\Y)=\frac{\pi(K)\pi(\Z|k)\pi(\Y|\Z)}{\pi(\Y)}.\nonumber
\end{align}
Now, applying an MCMC algorithm with $\pi(\Z,K|\Y)$ as the target density will give a (marginal) posterior of $K$, $\pi(K|\Y)$.

\subsubsection{Nested model}
\cite{peixoto17a} presented an example of a synthetic network containing 64 \textit{cliques} of 10 nodes. The nodes are connected with each other within a clique, while there is no edges between nodes in any two different cliques. The results of fitting the microcanonical SBM (Section \ref{sect.micro}) suggested an optimal of 32 groups, each containing two cliques, mean that the model suffers from \textit{underfitting} rather than \textit{overfitting}, due to the fact that the maximum detectable $K$ is (proportional to) $\sqrt{n}$ \citep{peixoto13}. Therefore, \cite{peixoto14b} proposed a nested SBM, also adopted by \cite{peixoto17a,peixoto17b}, which can be described using the aforementioned level-one graph $\mathcal{G}^{(1)}$. While $\mathcal{G}^{(0)}$ is characterised by the original SBM, the resulting $\mathcal{G}^{(1)}$ is characterised by another SBM, which results in $\mathcal{G}^{(2)}$, and so on, until there is only one group at the top level. Accompanied by the moves described above for modelling and inferring $K$, they found that this nested SBM managed to overcome the underfitting issue, while discovering a hierarchical structure.

\section{Comparison} \label{sect.compare_graph}
In this section, an overall comparison in two aspects is provided between the works reviewed. The first aspect concerns the performances of the models that all applied to two real-world examples. The second aspect concerns the approaches reviewed in Sections \ref{sect.graph_cluster}, \ref{sect.graph_inf} and \ref{sect.graph_K}.

\subsection{Real-world examples and performances} \label{sect.compare_eg}

A classic undirected network studied in the literature is the karate club reported by \cite{zachary77}. Due to the nature of the study, it was known that the club has been split into two main factions over a disagreement, eventually forming two smaller clubs. \cite{zachary77} managed to observe the friendship among $n=34$ of the members, and compiled the undirected network of $M=78$ edges, which is plotted in Figure \ref{fig.karate_plot} (there were more members who however did not interact with this biggest component of the network). It has been studied for both community detection methods and SBMs \citep{kn11,ysjkmzzz14,yan16,vv18,ls19b}. Within the community detection algorithms, \cite{gn02} and \cite{ng04} managed to reveal the two true groups, \cite{bgll08} stopped short at four groups. But this should not be taken as an indication of superiority automatically, as, in fact, there is also no unanimous agreement within the SBMs. On one hand, the models mostly agreed on the optimal $K$, as \cite{dkmz11} found $K=2$ using their belief propagation algorithm and the observed data log-likelihood criterion, and \cite{nr16} found that the posterior $\pi(K|\Y)$ in their DC-SBM is indeed maximised at $K=2$. On the other hand, while the DC-SBM \citep{kn11} revealed the two known factions that the original SBM failed, the likelihood ratio test by \cite{ysjkmzzz14} suggested that there were not enough evidence to suggest that the network was generated by DC-SBM (against the original SBM), even when the inhomogeneous degree distribution prompted the degree correction in the first place. Furthermore, \cite{yan16} found that using the ICL criterion selected $K=1$ under the DC-SBM, meaning that any blocking leads to overfitting, while $K=4$ was selected under the original SBM, with different groups corresponding to different levels of node degrees.
\begin{knitrout}
\definecolor{shadecolor}{rgb}{0.969, 0.969, 0.969}\color{fgcolor}\begin{figure}[h!]

{\centering \includegraphics[width=0.75\linewidth]{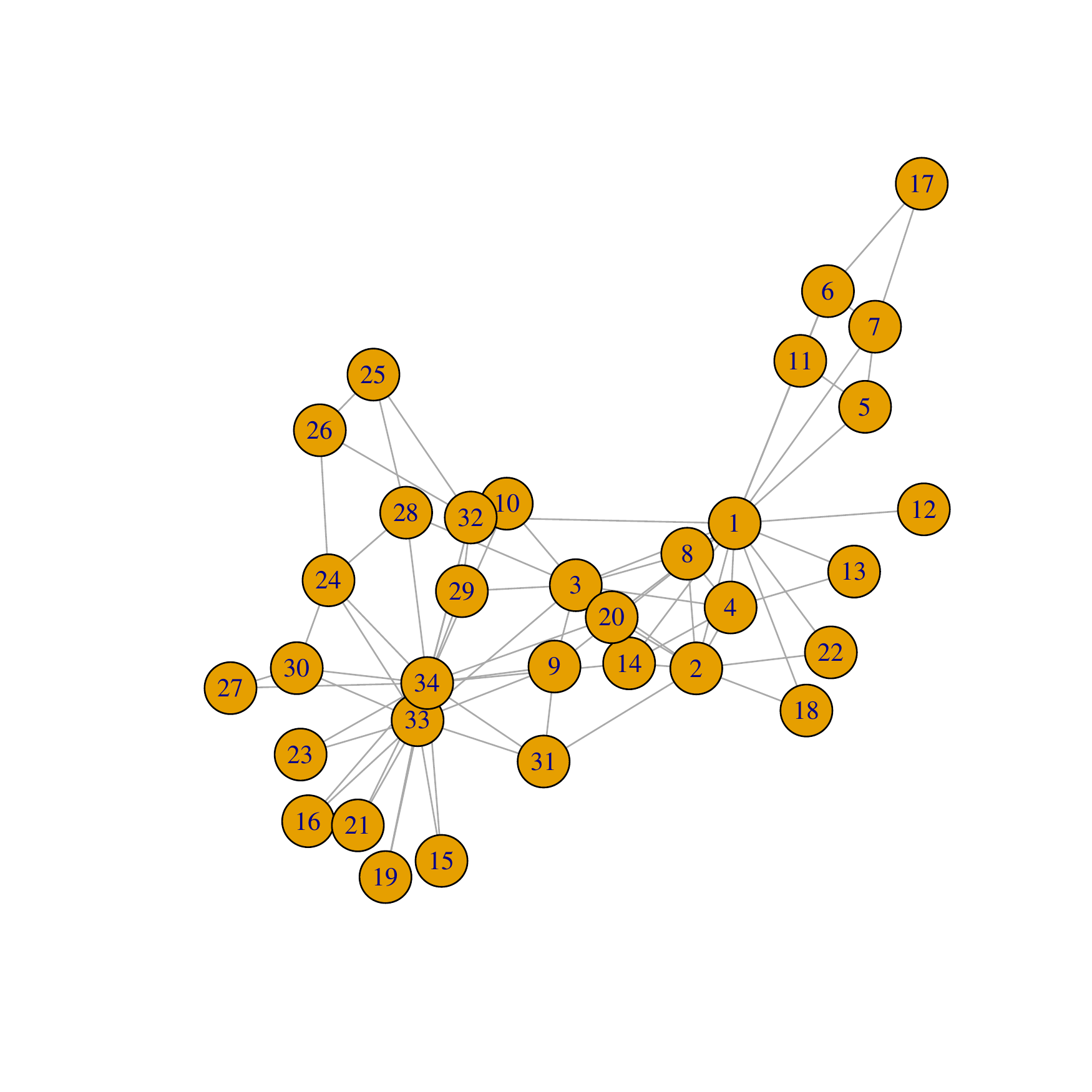} 

}

\caption[The karate club network by \cite{zachary77}]{The karate club network by \cite{zachary77}.}\label{fig.karate_plot}
\end{figure}

\end{knitrout}

Another example is the directed network of political blogs by \cite{ag05} during the 2004 U.S. Presidential Election. While it contains 1490 blogs in total, usually only the giant component of 1222 blogs (nodes) with 19089 is analysed. Furthermore, the blogs were labeled as either ``liberal'' or ``conservative'' by \cite{ag05} according to their political leanings, providing some sort of ground truth. By forcing $K=2$, the DC-SBM by \cite{kn11} again showed a high agreement with the ground truth, while the original SBM did not. The evidence for the former is supported by the likelihood ratio test by \cite{ysjkmzzz14}. \cite{yan16} presented a mixed picture in which the DC-SBM always dominated the original SBM, even more so at smaller $K$, but the ICL criterion actually increased with $K$, thus potentially requiring some penalty. While \cite{wb17} selected $K=4$ using their penalised log-likelihood, closer investigation revealed that one ground-truth group matched well with one inferred group, while the other ground-truth group is split into three smaller inferred groups. However, \cite{hqyz19} found that, at a certain value of the tuning parameter $\lambda$ (Section \ref{sect.criteria}) the penalised log-likelihood selected $K=1$, therefore arguing for their corrected BIC that has a heavier penalty. Finally, the nested SBM \citep{peixoto17a} managed to cluster the nodes beneath the two main political factions, and discovered 20 groups in the lowest-level SBM according to the DC-SBM. While there is an agreement over the big picture, the models usually differ in the fine details.

\begin{sidewaystable}[!htbp]
  \centering
  \begin{tabular}{|l|l|l|l|l|l|l|}
    \hline
    \textbf{Article} & \multicolumn{2}{c|}{\textbf{Type of graph}} & \textbf{SBM} & \textbf{Inference} & \textbf{Clustering} & \textbf{Number of groups}\\
    \hline
    \cite{sn97} & Binary & Undirected & Yes & MLE/EM/MCMC & Hard & Fixed (2)\\
    \cite{ns01} & Either & Either & Yes & MCMC & Hard & Criterion (Information)\\
    \cite{tallberg05} & Discrete & Directed & Yes & MCMC & Hard & Fixed (3)\\
    \cite{kks06} & Valued & Directed & Yes & Variational & Soft & Dirichlet process\\
    \cite{abfx08} & Binary & Directed & Yes & Variational & Soft & Criterion (BIC)\\
    \cite{dpr08} & Binary & Undirected & Yes & Variational & Hard & Criterion (ICL)\\
    \cite{mgj09} & Binary & Directed & No & MCMC & Hard & Indian buffet process\\
    \cite{dkmz11} & Binary & Directed & Yes & Belief propagation & Hard & Criterion (Observed data log-like.)\\
    \cite{kn11} & Valued & Undirected & Yes & Greedy algo.+MLE & Hard & Fixed (2)\\
    \cite{msh11} & Binary & Undirected & No & MCMC+HMC & Hard & Dirichlet process\\
    \cite{gmgfb12} & Binary & Undirected & Yes & Variational & Soft & Criterion (Perplexity)\\
    \cite{lba12} & Binary & Undirected & Yes & Variational & Hard & Criterion (Marginal log-like.)\\
%    \cite{gbr12,gbr18} & Binary & Directed & No & MAP & Hard & Estimated\\
    \cite{pkg12} & Binary & Undirected & No & MCMC & Hard & Dirichlet process\\
    \cite{kgbs13} & Binary & Undirected & Yes & Variational & Soft & Dirichlet process\\
    \cite{mmfh13} & Binary & Either & Yes & MCMC & Hard & Estimated\\
    \cite{vhs13} & Discrete & Either & Yes & Variational & Hard & Fixed (5, 20)\\
    \cite{ysjkmzzz14} & Valued & Undirected & Yes & EM & Hard & Criterion (Observed data log-like.)\\
    \cite{ajc15} & Valued & Directed & Yes & Variational & Hard & Criterion (Marginal log-like.)\\
    \cite{cl15} & Binary & Directed & Yes & Greedy algo. & Hard & Criterion (ICL)\\
    \cite{zhou15} & Valued & Undirected & No & MCMC & Hard & Hier. Gamma process\\
    \cite{ggms16} & Valued & Bipartite & Yes & Variational & Soft & Fixed (10)\\
    \cite{hkk16} & Binary & Directed & Yes & Variational & Hard & Criterion (F${}^2$IC)\\
    \cite{law16} & Binary & Undirected & Yes & MCMC & Soft & Criterion (Perplexity)\\
    \cite{nr16} & Binary & Undirected & Yes & MCMC & Hard & Estimated\\
    \cite{vmgs16} & Binary & Undirected & Yes & Unknown & Hard & Unknown\\
    \cite{yan16} & Binary & Undirected & Yes & Unknown & Hard & Criterion (BIC)\\
    \cite{peixoto15b,peixoto17a,peixoto17b,peixoto18a} & Either & Either & Yes & MCMC & Either & Estimated\\
    \cite{rvw17} & Binary & Bipartite & No & MCMC & Soft & Criterion (DIC)\\
    \cite{wb17} & Binary & Undirected & Yes & Variational & Hard & Criterion (Penalised log-like.)\\
    \cite{gpa18} & Valued & Bipartite & Yes & MCMC & Hard & Estimated\\
    \cite{vv18} & Binary & Undirected & Yes & Unknown & Hard & Fixed\\
    \cite{hqyz19} & Binary & Undirected & Yes & Unknown & Hard & Criterion (BIC, corrected)\\
    \cite{ls19b} & Valued & Undirected & Yes & MCMC & Hard & Fixed (2)\\
    \cite{sh19} & Binary & Undirected & Yes & MCMC & Hard & Estimated\\
    \cite{sbknm19} & Binary & Undirected & Yes & Variational & Hard & Criterion (ICL)\\
%    \hline
%    \cite{hfx10} & Valued & Directed & No & MLE & Yes & N/A \\
%    \cite{kh14} & Valued & Either & No & MLE & Yes & N/A \\
    \hline
    \end{tabular}
  \caption{Models for graphs without longitudinal or topic modelling.}
  \label{table.graph}
\end{sidewaystable}

\subsection{Approaches}
Several aspects of SBMs and related models have been considered so far, namely the type of graph, whether the model is an SBM or not, the inference approach, the clustering approach, the number of groups, and whether there is longitudinal modelling. However, these aspects are not completely independent of each other. For example, opting for a latent feature model \citep{mgj09,msh11} not only deviates from an SBM, but also increases the computation complexity because the number of latent variable combination increases from $K^n$ to $2^{Kn}$, thus in turn prompting for an efficient and scalable inference algorithm. Another example is that a soft clustering approach influences partly how $K$ is being modelled or selected. The models considered here either opted for modelling by a Dirichlet process \citep{kks06,kgbs13,fcx15} or selecting by a criterion \citep{abfx08,fsx09,xfs10,gmgfb12,law16}. Also, the added computational complexity prompted \cite{law16} to propose an inference algorithm which exploits the sparity of graphs. Finally, using a variational approach more often results in $K$ being selected by a criterion \citep{abfx08,fsx09,xfs10,gmgfb12,lba12,ajc15,mrv15,hkk16,mm17} than not \citep{kks06,kgbs13,vhs13}. Note that these are not intended as definitive arguments but to highlight that the modelling, inference, and clustering approaches, and the issue with the number of groups, are quite interconnected. For a comprehensive comparison, please see Tables \ref{table.graph} and \ref{table.longitudinal} for models with and without longitudinal modelling, respectively.

\section{SBM with longitudinal modelling} \label{sect.graph_longitudinal}
The SBMs and related models introduced so far assume that the graph is observed at one instant, which can be regarded as a cross-section of a graph that is evolving over time. If temporal information of the interactions is available, or the graph is observed at multiple instants, longitudinal or dynamic models for graphs can be applied. Therefore, articles that incoporate longitudinal modelling will be reviewed in this section, providing a partial answer to question \ref{q.extension} in Section \ref{sect.intro}.

In the dynamic model by \cite{fsx09} and \cite{xfs10}, which are direct extensions of the MMSBM by \cite{abfx08}, the membership probabilities of node $p$ is now indexed by time $t$, denoted by $\t[p]^{(t)}$, and is dependent on $\t[p]^{(t-1)}$ via a state space model. Similarly, the block matrix now becomes $\C^{(t)}$, and evolves over time according to a separate and independent state space model. The latent pairwise group memberships at time $t$, $\Z^{(t)}$, is then generated by $\T^{(t)}=\left(\t[1]^{(t)}~\t[2]^{(t)}~\cdots~\t[n]^{(t)}\right)^T$. Finally, the observed graph $\Y^{(t)}$ is assumed to arise from an MMSBM \citep{abfx08} with parameters $\C^{(t)}$ and latent variables $\Z^{(t)}$. 

\cite{fcx15} proposed two dynamic models, by combining the Dirichlet process and the MMSBM. In their mixture time variant (MTV) model, at each instant the group memberships $\Z^{(t)}$ are drawn according to the membership probability vector $\T^{(t)}$, elements of which arises from a Dirichlet process. The process parameters in turn depend on the history of group memberships $\{\Z^{(1)},\Z^{(2)},\ldots,\Z^{(t-1)}\}$. In the mixture time invariant (MTI) model, at each instant the group memberships $\Z[p]^{(t)}$ depends on $K$ time-invariant $\t[p]^{(i)}(i=1,2,\ldots,K)$, also from a Dirichlet process, \textit{and} the previous group memberships $\Z^{(t-1)}$. The graph $\Y^{(t)}$ is then generated from an MMSBM with $\Z^{(t)}$, and a universal block matrix $\C$. This means that the historical group memberships influence the distribution of the current ones, in the MTV and MTI models, through the Dirichlet process parameters and the group-specific (and time-invariant) membership probabilities, respectively.

\cite{ty11,ty14} adopted a similar idea in their dynamic SBM with temporal Dirichlet process. As hard clustering is used here, there is one membership vector $\t^{(t)}$ for all nodes at each instant. However, this $\t^{(t)}$ is dependent on the Dirichlet process parameter \textit{and} the group memberships $\Z^{(t-1)}$ at the previous instant, so that the collection $\{\t^{(t)}\}$ arises from what they called the recurrent Chinese restaurant process. The ingredients for generating the graph $\Y^{(t)}$ include the block matrix $\C$, which is constant over time, and the group memberships $\Z^{(t)}$, which are generated by $\t^{(t)}$ in turn.

\cite{yczgj11} considered a simpler evolution mechanism of the group memberships in their dynamic SBM. A Markov chain is assumed for $\Z^{(t)}$, which means, for node $p$, $\Z[p]^{(t-1)}$ moves to $\Z[p]^{(t)}$ according to a transition matrix, which remains unchanged over time. Also assumed constant over time are the model parameters $\t$ and $\C$. \cite{mm17} proposed a similar model, except that the graph is allowed to be valued, in which case each element in $\C$ does not necessarily mean the edge probability. They noted that allowing the group memberships \textit{and} the block matrix to vary over time will lead to identifiability and label-switching issues, therefore fixing $\C$ to be constant over time.

\cite{xh13} used a dynamic SBM quite different to the ones introduced so far. They considered the observed density of edges between two groups to be a noisy observation of a dynamic system, in which the corresponding element in the block matrix is the state. While $\C^{(t)}$ is modelled by a state space model, there is none specified for the time-specific group memberships $\Z^{(t)}$.

Another distinctive model is the autoregressive SBM \citep{len17}, in which the group membership $\Z[p]^{(t)}$ follows a continuous-time Markov chain (CTMC), meaning that node $p$ spends an exponentially distributed time in a group, before moving to another group chosen uniformly at random from the remaining groups. If dyad $(p,q)$ \textit{belongs completely} to one group, that is, both $p$ and $q$ belong to the same group, the presence or absence of an edge over time is modelled by a separate CTMC, the transition rates of which are universal to all dyads that belong completely to the same group as $(p,q)$. For all the remaining dyads where the group memberships do not coincide, there is one extra set of transition rates governing the independent edge process for each dyad. Essentially, there are $n(n-1)/2$ CTMCs modelling the dyads, the parameters of which are determined, as always in SBMs, by the group memberships.

Whilst also using a CTMC in their dynamic model, \cite{zmn17} placed a it on the edge existence/absence $\Y[pq]^{(t)}$, instead of the group membership $\Z[p]$. Using the DC-SBM \eqref{eqn.graph_ypq_kn11}, the rate that edges are removed depends on the group memberships of $p$ and $q$, while the rate that edges are added is the product of the edge-removal rate and $\phi_p\phi_q\Z[p]^T\C\Z[q]$, which is the same term used in the static model \eqref{eqn.graph_ypq_kn11}. Using the Markov property of a CTMC, they worked out the probabilities of the graph observed at one time point, \textit{given} the graph at the previous time point, thus the whole likelihood. For inference, \textit{conditional on $\Z$}, the MLEs of the other parameters can be obtained analytically (in terms of $\Z$). These estimates are then plugged in to the original log-likelihood to arrive at the profile log-likelihood, which is then maximised using a greedy algorithm, to obtain the MLEs of $\Z$.

\replaced{\cite{peixoto15a} proposed two versions of modelling longitudinal or temporal networks, in which group memberships are the same across layers or time, based on the MDL approach \citep{peixoto17a}. In the first, the total numbers of edges between the groups across \textit{all} layers are viewed as a \textit{collapsed} graph that arises from one SBM (according to, for example, \eqref{eqn.graph_lik_hard_re} or \eqref{eqn.graph_ypq_poisson_ij}). The edges between groups $i$ and $j$ are then assumed to be uniformly distributed, conditioned only on the number of edges between $i$ and $j$ \textit{in each layer}. In the second version, each layer is generated from an independent SBM, but nodes are allowed to belong only to a subset of the layers. A node not belonging to a layer means that the node is forbidden to have edges in that layer. Equivalence can be established between special cases of the two versions for non-degree-corrected SBMs, but their degree-corrected counterparts are in general not equivalent. This is because degree variability across layers is allowed for the second version but not the first. For temporal modelling, unlike the models introduced above, there are no temporal dynamics explicitly specified. However, \cite{peixoto15a} suggested binning the times so that each bin can be viewed as a layer, within which the large-scale network structure does not change significantly. Depending on the version chosen and whether degree correction is incorporated, how the edges can be assigned to the nodes at different layers is flexibly regulated. Finally, selecting the most appropriate time binning, alongside the choice between the two versions and that of degree correction or not, can be aided by model selection through the use of the MDL.}{}

The models introduced so far model the evolution of a graph, usually a binary one, and observed at different instants. They are however not suitable for recurrent interaction events in a network, of which the times of interaction are random in nature. In the aforementioned \cite{dbs13}, the interactions for the dyad $(p,q)$ arise from a Poisson process with intensity $\exp(\Z[p]^T\C\Z[q])$. The intensity can be extended to depend on not only the group memberships only but also on the historical interactions. Specifically, it is the dot product of a vector summarising the interactions of the dyad and $\Z[p]^T\C\Z[q]$, which is now allowed to be vector-valued (essentially making $\C$ a 3-dimensional array). However, due to such a generalisation, it is misleading to continue calling such model a Poisson process, because the piecewise constant intensity depends on the history of the process, thus making it self-exciting in nature.

\cite{mrv15} formulated a similar model using conditional Poisson processes. The block matrix $\C(t)$ is now a $K\times{}K$ matrix of intensity \textit{functions}, not probabilities or scalar parameters. \textit{Conditional on $\Z[p]$ and $\Z[q]$}, which are assumed to remain unchanged over the course of time, the interactions of the dyad $(p,q)$ follows a nonhomogeneous Poisson process with intensity $\Z[p]^T\C(t)\Z[q]$. Similar to \cite{mm17}, identifiability and label-switching issues were also discussed.

It should be noted that having both the temporal information in the data and a longitudinal/dynamic SBM does not guarantee that the actual groups will be discovered. \cite{gzcmp16} studied a dynamic SBM, and derived a threshold as a function of the rate of change and the strength of the groups/communities. They found that, below that threshold and empirically, no efficient inference algorithms can identify the groups better than chance.

\section{Topic models} \label{sect.topic}
In this section, we briefly introduce the general form of topic models, in particular latent Dirichlet allocation, to prepare for Section \ref{sect.both}, where they are incoporated in SBMs. While topic models are not the main focus of this review, we will discuss aspects including the inference approach, dealing with the number of topics, and longitudinal modelling, which have been covered in Sections \ref{sect.graph_inf}, \ref{sect.graph_K}, \ref{sect.graph_longitudinal} for SBMs, respectively.

The main goal of topic modelling is to cluster a collection of \textit{documents} into different \textit{topics}. Each latent topic is represented by its distribution over the words that appear in the documents, with such distribution usually being visualised by wordclouds. The roles of the documents, the topics, and the words are then analogous to those of the nodes, the groups, and the edges, respectively. One major difference is that a document is usually assumed to be independent of other documents \textit{apriori}, whereas a node in a graph is defined by its interactions, or the lack thereof, with others. Another difference is that soft clustering is the dominating approach in topic modelling, meaning that each document usually has non-binary weights over multiple topics. For example, an article in genetics may be about biology, chemistry, statistics, and medicine, for 40\%, 30\%, 20\% and 10\%, respectively. This is in contrast to the (usual) hard clustering approach in SBMs.

We first introduce some terminology and notation, which may look different to that in the literature, but is intended to align with what we have introduced for SBMs. Assume that the there are $m$ documents, and $n$ distinct words in total, denoted by $\V:=\left(\V[1]~\V[2]~\cdots~\V[n]\right)^T$. In document $p=1,2,\ldots,m$, there are $N_p$ words. The $q$-th word is represented by an $n$-vector $\W[pq]:=\left(\W[pq1]~\W[pq2]~\ldots~\W[pqn]\right)^T$, in which one element is 1, the rest 0. If $\W[pqk]=1$, this means the $k$-th word in $\V$ is the one used as the $q$-th word in document $p$. We also define the $N_p\times{}n$ matrix $\W[p]:=\left(\W[p1]~\W[p2]~\cdots~\W[pN_p]\right)^T$, and  $\W:=\{\W[1]~\W[2]~\cdots~\W[m]\}$, which essentially is the whole of the data. Finally, an $m\times{}n$ matrix $\mathbf{M}$, called the document-word frequency matrix, is defined. Each element $\mathbf{M}_{pk}~(p=1,2,\ldots,m;k=1,2,\ldots,n)$ represents the frequency of the word $\V[k]$ in the $p$-th document. 

A common assumption in topic modelling is that there are $K$ latent \textit{topics}. Associated with the $i$-th topic is an $n$-vector $\p[i]:=\left(\p[i1]~\p[i2]~\cdots~\p[in]\right)^T$, subject to the constraint $\p[i]^T\boldsymbol{1}_n=1$, which represents the distribution of the vocabulary in this topic. The sequence $\{\p[i]\}~(i=1,2,\ldots,K)$ is assumed to be independent and identically distributed (i.i.d.) according to the Dirichlet$(\beta\boldsymbol{1}_K)$ distribution. Collectively, we write $\Phi:=\left(\p[1]~\p[2]~\cdots~\p[K]\right)^T$, which is a $K\times{}n$ matrix. 

In a similar fashion, associated with the $p$-th document is a $K$-vector $\t[p]$, subject to the constraint $\t[p]^T\boldsymbol{1}_K=1$, which represents the distribution of the $K$ topics for this document, or its mixed membership in the topics. The sequence $\{\t[p]\}~(p=1,2,\ldots,m)$ is assumed to be i.i.d. according to the Dirichlet$(\a\boldsymbol{1}_K)$ distribution. Collectively, we write $\T:=\left(\t[1]~\t[2]~\cdots~\t[m]\right)^T$, which is an $m\times{}K$ matrix. This is similar to the definitions in Section \ref{sect.graph_cluster}.

The main difference between various topic models is the generating mechanism of the words in the documents. The principle in latent Dirichlet allocation (LDA) by \citep{bnj03} is that a document can belong to different topics when generating each word. Specifically, for document $p$, associated with the $q$-th word $(q=1,2,\ldots,N_p)$ is a $K$-vector latent variable, denoted by $\Z[pq]$. Only one element of $\Z[pq]$ is 1, representing the topic the document belongs to \textit{for this particular word}, the rest of which 0. (This is similar to the latent group membership of a node $p$ \textit{for its particular interaction with $q$} in MMSBM \citep{abfx08}.) Collectively, we write $\Z[p]:=\left(\Z[p1]~\Z[p2]~\cdots~\Z[pN_p]\right)^T$, which is an $N_p\times{}K$ matrix, and $\Z:=\{\Z[1],\Z[2],\ldots,\Z[n]\}$, which is a sequence of matrices as well as a collection of all latent variables.

%To calculate the likelihood, we first break it down into components. The probability that topic $i$ is chosen for word $q$ in document $p$ is given by $\Pr(\Z[pqi]=1)=\T[pi]$, while the probability that $\V[k]$ is selected for this particular word, given topic $i$ is chosen, is $\Pr\left(\W[pqk]=1|\Z[pqi]=1\right)=\P[ik]$. The likelihood contribution by document $p$, given $\Z[p]$, is
%\begin{align}
%  \pi\left(\W[p]|\Z[p],\P\right)=\Prod[N_p]{q=1}\pi\left(\W[pq]|\Z[pq],\P\right)=\Prod[N_p]{q=1}\Z[pq]^T\P\W[pq],\label{eqn.topic_lik_indiv}
%\end{align}
%and the likelihood is multiplying \eqref{eqn.topic_lik_indiv} over $p$. The expression is made delibrately close to that for SBMs, such as \eqref{eqn.graph_lik_hard}. 

Now, with the data $\W$, the latent variables $\Z$, the parameters $\T$ and $\P$, the likelihood can be computed in the usual manner, and inference carried out. As with MMSBM (Section \ref{sect.graph_cluster}), the main goal of inference is for $\T$ (and $\P$) but not $\Z$. Instead of going through the derivations of the posterior in detail, we shall mention works which have worked on various aspects in Section \ref{sect.graph_inf}, \ref{sect.graph_K} and \ref{sect.graph_longitudinal}.

\subsection{Inference approach}\label{sect.topic_inf}
In the usual approach which is Bayesian inference, in addition to the Dirichlet distribution assumptions made for $\P$ and $\T$, once the priors are assigned to $\a$ and $\b$, the joint posterior of $\Z$, $\P$, $\T$, $\a$ and $\b$ can be derived.
%\begin{align}
%  &\pi\left(\Z,\P,\T,\a,\b|\W\right)\propto\pi\left(\W,\Z,\P,\T,\a,\b\right)\nonumber\\
%  &\quad=\pi\left(\W|\Z,\P\right)\times\pi\left(\Z|\T\right)\times\pi\left(\P|\b\right)\times\pi\left(\T|\a\right)\times\pi\left(\a\right)\times\pi\left(\b\right)\nonumber\\
%  &\quad=\Prod[m]{p=1}\Prod[N_p]{q=1}\big[\pi\left(\W[pq]|\Z[pq],\P\right)\pi\left(\Z[pq]|\T\right)\big]\Prod[K]{i=1}\pi\left(\p[i]|\b\right)\Prod[m]{p=1}\pi\left(\t[p]|\a\right)\times\pi\left(\a\right)\pi\left(\b\right)\nonumber\\
%  \begin{split}
%    &\quad=\Prod[m]{p=1}\Prod[N_p]{q=1}\big[\left(\Z[pq]^T\P\W[pq]\right)\times\left(\Z[pq]^T\t[p]\right)\big]\\
%    &\qquad\times\Prod[K]{i=1}\left[\Gamma\left(n\b\right)\one{\p[i]^T\boldsymbol{1}_n=1}\Prod[n]{k=1}\frac{\P[iq]^{\b-1}}{\Gamma\left(\b\right)}\right]\times\b^{c-1}e^{-d\b}\\
%    &\qquad\times\Prod[m]{p=1}\left[\Gamma\left(K\a\right)\one{\t[p]^T\boldsymbol{1}_K=1}\Prod[K]{i=1}\frac{\T[pi]^{\a-1}}{\Gamma\left(\a\right)}\right]\times\a^{a-1}e^{-b\a}.
%  \end{split}\label{eqn.topic_inf_joint}
%\end{align}
Unlike in SBMs, even with the soft clustering approach, the computational complexity grows linearly, but not quadratically, with the number of documents $m$.

Similar to what has been described in the beginning of Section \ref{sect.graph_inf}, because of the use of conjugate priors for $\P$, $\T$, $\a$ and $\b$, for algorithmic simplicity a Gibbs sampler can be derived, in which the parameters and the latent variables can be updated via individual Gibbs steps. However, \cite{gs04b} were not interested in $\P$ and $\T$ and therefore integrated them out, to obtain a \textit{collapsed} Gibbs sampler. Furthermore, as they fixed the values for $\a$ and $\b$, only the latent groups $\Z$ are to be inferred. This is similar to collapsing the hard clustering SBMs in, for example, \cite{mmfh13}, \cite{cl15}, and \cite{peixoto17a,peixoto17b}, although the parameters $\C$ in the soft clustering MMSBM \citep{abfx08} are not being integrated out. Other uses of MCMC algorithms include \cite{tjbb05,rdc08,pr08,ax10} and \cite{gg11}. However, such algorithms are more useful when the number of topics $K$ is being modelled (Section \ref{sect.topic_K}), rather than selected by a criterion, by processes such as the hierarchical Dirichlet process and the Indian buffet process.

\cite{bnj03}, along with \cite{wpb11} used variational inference, the general formulation of which is described in Section \ref{sect.graph_inf_variational}. \cite{bl06} combined variational inference and a Kalman filter for their dynamic topic model, in which a state space model is incorporated, to be explained in Section \ref{sect.topic_longitudinal}. \cite{asmm10} focussed on finding the optimal number of topics and did not explicitly state their inference approach.

\subsection{Number of topics}\label{sect.topic_K}
Using the same notation $K$ as in Section \ref{sect.graph} is because the number of topics in a topic model is analogous to the number of groups in an SBM. Therefore, there also exists the issue of whether $K$ should be fixed, selected by a criterion, or modelled. Of the topic models reviewed here, only \cite{bl06} used a fixed number of topics ($K=20$). \cite{bnj03} used the perplexity, which is the likelihood of held-out test data, to determine $K$. This is (almost) the same as the definition of perplexity in SBMs \citep{gmgfb12,dbs13,law16}. \cite{gs04b} selected $K$ by the marginal log-likelihood, which is similar to \cite{lba12} for SBMs, although they have not provided the derivations of this quantity $\log\pi\left(\W\right)$, or approximations thereof. 

\cite{asmm10} proposed an empirical measure to find the optimal $K$. They started with viewing that LDA essentially ``factorises'' the document-word frequency matrix $\mathbf{M}$ into the two stochastic matrices $\P$ and $\T$. (Of course, this does not mean that $\mathbf{M}=\T\P$, however.) By defining the vector of document lengths $\boldsymbol{N}:=\left(N_1~N_2~\cdots~N_m\right)^T$, they gave the expression of the criterion for selecting $K$, as the symmetric K-L divergence between the distribution of the singular values of $\P$, and the distribution obtained by normalising $\boldsymbol{N}^T\T$.

\subsubsection{Hierarchical Dirichlet process} \label{sect.topic_K_hdp}
The Dirichlet process (DP) and its hierarchical version have to be introduced before incorporating into topic models. As it is only used in some of the models reviewed, we focus on the ideas behind these \textit{nonparametric Bayesian} processes, rather than their technical details.

The topic model introduced in the beginning of this section assumes that, for each document, there is a same number ($K$) of topics to choose from, and the membership probabilities $\t[p]~(p=1,2,\ldots,m)$ form a $K$-vector which comes from a Dirichlet$(\alpha\boldsymbol{1}_K)$ \textit{distribution}. Dirichlet process takes this one step further by not requiring $K$ to be pre-specified. While the elements of $\t[p]$ drawn from a DP still sum to 1, $K$ varies between i.i.d. samples and is implied by the length of $\t[p]$. This means, for example, $\t[p]=(0.2~0.5~0.3)^T$ and $\t[p]=(0.15~0.45~0.35~0.05)^T$ can be two independent samples from the same DP, implying $K=3$ and $K=4$, respectively.

To take the DP one step further, \cite{tjbb05} assumed that $\t[p]$ comes from a DP, the parameters of which come from another DP. Introducting this structure in their hierarchical Dirichlet process (HDP) leads to a clustering effect. In the context of topic models, the topics generated for each document are uncorrelated in a DP, while the same set of topics is shared across all documents in an HDP.

This description of the DP and HDP is an informal one, and has omitted the technical details of actual sampling from a DP; see, for example, \cite{tjbb05,rdc08}, and \cite{ax10}. What is to be highlighted here is that $K$ arises naturally when sampling from a DP or HDP, and therefore does not need to be pre-specified. As these nonparametric Bayesian processes concerns the generation of the mixing vectors $\t[p]$ and is independent of the topic models, it can be incorporated in models that require these membership probabilities, hence its use in the SBMs discussed in Section \ref{sect.graph_K_modelled}.

It has been argued in Section \ref{sect.graph_inf} and shown by, for example, \cite{gs04b} that a Gibbs sampler can be derived for the parameters and the latent variables $\Z$ of the topic model when $K$ is fixed. \cite{tjbb05} have shown that a Gibbs sampler is also possible under the HDP formulation, hence the preference to using MCMC as the inference algorithm too, in \cite{rdc08} and \cite{ax10}. While \cite{pr08} proposed two MCMC algorithms to improve the then existing ones for HDP models in general, \cite{wpb11} pointed out the limitation that such algorithms require multiple passes through all the data and are therefore not very scalable. They proposed a variational infernce algorithm as an alternative.

So far, for a topic model or SBM, incorporating an HDP means that the quantity $K$ corresponds to is potentially infinite and to be modelled. The equivalent for latent feature models exists and is termed the Indian buffet process, and again a Gibbs sampler for such models is possible \citep{mgj09}. For a detailed introduction and an extensive review, see \cite{gg11}.

\subsection{Longitudinal modelling}\label{sect.topic_longitudinal}
As temporal information is sometimes available for text data, such as academic articles, longitudinal or dynamic models can be incorporated in a similar way as they are for SBMs or other models for graphs. \cite{bl06} proposed a dynamic topic model, in which the distribution of vocabulary for the $i$-th topic is now indexed by time $t$, denoted by $\p[i]^{(t)}$, and is dependent on $\p[i]^{(t-1)}$ via a state space model. The incorporation of a state space model is similar to that in \cite{fsx09,xfs10} and \cite{xh13}, mentioned in Section \ref{sect.graph_longitudinal}. The main difference is the generation the membership probabilities, as once the $p$-th document is generated from $\t[p]$ there is no need to evolve its distribution over the topics. Instead, there is a ``universal'' vector of membership probabilities, denoted by $\boldsymbol{\psi}^{(t)}$, and depends on $\boldsymbol{\psi}^{(t-1)}$ via a separate state space model. Assuming that, without loss of generality, the $p$-th document is to occur at time $t$, $\t[p]$ is generated from $\boldsymbol{\psi}^{(t)}$ with random noise. Words of the document are then generated in the usual way according to $\t[p]$ and $\p^{(t)}$.

As mentioned before, in a non-dynamic model that models $K$, the memberships $\t[p]$ arises from an HDP. \cite{rdc08} introduced temporal dependency in their dynamic model, by assuming that the documents occurred sequentially, and that $\t[p]$ depends on both $\t[p-1]$ and an HDP which represents innovation. \cite{ax10} modified the HDP in a slightly different way. The membership probabilities now come from, loosely speaking, an evolving DP, which depends on the process at the previous time point through a state space model. They argued that their dynamic model allows topics to be born and die at any instant.

\section{SBM with topic modelling} \label{sect.both}
In this section articles that combine a model for graphs and a topic model are reviewed. Similar to Section \ref{sect.topic}, different aspects related to modelling and inference approaches are discussed, thus providing answers to questions \ref{q.extension} and \ref{q.extension_aspects} in Section \ref{sect.intro}.

Before looking at SBMs with topic modelling for graph \textit{and} textual data, it is useful to look at a recently proposed SBM \textit{for} textual data by \cite{gpa18}. Instead of probabilitistically factorising the document-word frequency matrix $M$ (Section \ref{sect.topic}), they treated the relations between the words and the documents as a bipartite graph, in which both the words and the documents are the nodes. By doing so, the SBMs reviewed before can be direct applied to the graph. In particular, they fit a nested SBM \citep{peixoto14b}, thus constructing a hierarchical structure and allowing the number of groups to be inferred. The proposed model was found to perform better than the traditional approach of LDA. It can be compared with the model by \cite{ggms16} for recommender systems, who also applied an SBM for seemingly non-relational data that can however be represented as graphs.

\subsection{Type of data and modelling approach} \label{sect.both_modelling}
In the articles reviewed in this section, the modelling approach is largely influenced by the type of data available, in particular whether the textual information is for the edges or nodes in the graph. The availability of such textual information is in turn determined by the nature of the data set itself. Therefore, we will review these two aspects together.

\subsubsection{Textual edges}
One famous example of data that contains both network and text information is the Enron Corpus, a large database of over 0.6 million emails by 158 employees of the Enron Corporation before the collapse of the company in 2001. The nature of an email network leads to a directed and valued graph, in which the nodes and edges are the employees and the email exchanges, respectively, with the latter of which the texts are associated. In terms of the models reviewed in Section \ref{sect.graph}, it has been studied by \cite{fsx09,xfs10,gmgfb12,dbs13,xh13} and \cite{mrv15}. Models in which both the graph and the texts are studied include \cite{zmlgz06,mwc07,pdbe08,scfs12,blz16} and \cite{cblr18}, which will be discussed individually.

\cite{zmlgz06} proposed two models in which the graph is not modelled explicitly. Instead, each email, or communication document in general, is generated by an extension of LDA. In the first model, the membership probabilities $\t$ is not associated with the document, but with the users involved with this document (for example, sender and receiver of an email). Then, the latent topic variable for word $q$ in document $p$, denoted by $\Z[pq]$ as in Section \ref{sect.topic}, is generated according to the membership probabilities of the users. There is an additional layer in which users come from different groups (communities), but it is not that each user has a mixed membership over the groups, but that for each group there is a distribution of users representing their participation. In the second model, associated with each group is the membership probabilities $\t$ of the topics, and associated with each topic is a distribution of the \textit{users}. This time, it is not made clear how a word in the document is generated from the user, who is in turn generated by the topic.

Similar to those by \cite{zmlgz06} is the author-recipient-topic model by \cite{mwc07}, in which the membership probabilities $\t$ are now specific to each author-recipient pair. For document $p$, the author and the set of recipients are treated as given. For word $q$ in this document, a recipient is selected at random uniformly, and the latent topic $\Z[pq]$ follows a multinomial distribution according to the aforementioned pair-specific membership probabilities. Finally, the word is generated, as usual, according to the word distribution $\p$ over the selected latent topic. \cite{pdbe08} extended the author-recipient-topic model, by adding the group element of the authors and recipients. In particular, chosen at random uniformly is not the recipient but the group. Then, similar to \cite{zmlgz06}, the authors and recipients are selected according to a group-specific distribution of users.

\cite{scfs12} introduced a topic user community model, which is different to those by \cite{zmlgz06} despite the similarities in the model names. For each sender, there is a group distribution, and for each sender-group pair, there is a distribution over the topics. There is one latent group for each document by the sender, and the latent topic variables for individual words are generated according to the aforementioned sender-group-specific probabilities. To complete the model specification, the recipients are chosen by a group-specific distribution of the latent group.

\cite{blz16} combined the SBM and LDA to form the stochastic topic block model (STBM). As in the SBM introduced in Section \ref{sect.graph}, the groups of nodes $p$ and $q$ arise from a multinomial distribution, and the edge variable $\Y[pq]$ follows a Bernoulli distribution with parameter $\Z[p]^T\C\Z[q]$, where $\Z[p]$ and $\Z[q]$ represent the memberships in the \textit{groups}, not the topics. Now, for each pair of groups $i$ and $j$, there is a specific vector $\t[ij]$ representing the memberships in the \textit{topics}, the collection of which, still denoted by $\T$, is a 3-dimensional array. Then, the latent topic variables for individual words of a document from $p$ to $q$ follow a multinomial distribution with parameters $\Z[p]^T\T\Z[q]$, \textit{conditional on} $\Y[pq]=1$. This means that the latent groups influence both the dyad and potentially the words in the document, if an edge exists.

\cite{cblr18} extended the STBM by incorporating a dynamic component, and is the only work reviewed in this section with longitudinal modelling, possibly due to complexity of combining a model for a graph and a topic model. Instead of conditioning on $\Y[pq]=1$, the occurrences of the documents from $p$ to $q$ arise from a nonhomogeneous Poisson process, according to the collection of group-pair-specific intensity functions and the latent groups $\Z[p]$ and $\Z[q]$. Furthermore, these intensity functions are piecewise constant, that is, constant within each of the time clusters, which are also latent variables and have to be inferred. Within each time cluster and conditioned on the existence of a document, the latent topic variables and individual words are again generated in the same way as in \cite{blz16}. Essentially, \cite{cblr18} proposed a model for simultaneously clustering three aspects, namely the nodes into groups, the documents into topics, and the occurrences into time clusters. It can also been seen as a generalisation, or even a direct combination, of both the STBM and the dynamic extension of the LDA.

\subsubsection{Textual nodes}
In another set of models, the entities are usually documents with texts and links between them. Note that a document is defined in its broad sense, as it can be a Wikipedia page, a blog post, or an academic article, the links in the last of which are citations or references. The data structure is then a graph, usually directed, with texts in the \textit{nodes}. We shall discuss a few of them.

\cite{lng09} argued that links between documents are not only determined by content similarity, but also by the connections between authors, because authors are naturally more aware of documents in their community and might not be aware of the possibly more relevant documents outside it. They introduced a topic-link LDA model, in which the edge probability between document $p$ and $q$ depends on a linear combination of document similarity and author similarity. The former similarity is the dot product of the topic memberships $\t[p]$ and $\t[q]$ of the documents, generated as described in Section \ref{sect.topic}, while the latter is the dot product of the memberships of the authors, drawn in a similar way from a separate Dirichlet distribution. The coauthorship network of the authors, however, is not incorporated to enrich the information on the author memberships.

\cite{cb10} proposed a relational topic model, in which the graph of the documents depends on their content similarity. First, each document is generated according to LDA. Next, for documents $p$ and $q$, their edge probability depends on the similarity between the latent variables $\Z[p]$ and $\Z[q]$ as defined in Section \ref{sect.topic}, quantified by a link probability function. Different version of this function are considered.

\cite{hex12} introduced a model called TopicBlock, in which a latent hierarchical or tree structure is assumed to generate the documents, which are the leaf nodes. Also defined is a hierarchical node $h$, that is, the root or anything between the root and the leaves. Associated with $h$ is a word distribution, denoted by $\p[h]$ for alignment with notation in Section \ref{sect.topic}, as well as a parameter for the edge probability between any two documents that share $h$ as their deepest common ancestor. Three things are then generated for each document. First, the path from the root to the document arises from a nested Chinese Restaurant process, which is related to the DP. Second, the words are generated according to a mixture of the distributions $\p[h]$ of all the nodes $h$ along the path. Finally, the presence or absence of edge of this document with another document follows a Bernoulli distribution according to the parameter associated with their deepest common ancestor. The use of a hierarchical latent structure to infer the number of groups or topics and for scalability is similar to \cite{peixoto17a}.

\subsection{Inference approach}
As it is illustrated in Sections \ref{sect.graph} and \ref{sect.topic} that a naive MCMC algorithm can be derived for a block model and a topic model, respectively, it is natural and possible to derive a similar algorithm for a model that combines both, if scalability is less of a concern compared to algorithmic simplicity. Each of \cite{zmlgz06,mwc07,pdbe08,hex12} and \cite{scfs12} have derived a Gibbs sampler for their corresponding model. In particular, \cite{hex12} followed \cite{gs04b} and integrated out the parameters other than the latent variables in their collapsed Gibbs sampler. 

On the other hand, the VEM algorithm described in Section \ref{sect.graph_inf_variational} is very general and equally feasible as the alternative. Therefore, it has been used by \cite{lng09,cb10,blz16} and \cite{cblr18} as the inference algorithm. \cite{blz16} and \cite{cblr18} observed that, in their STBM and dynamic counterpart, the equivalent of the lower bound in \eqref{eqn.graph_inf_lower_bound} can be split into two components, one depending on the variational distribution of the latent topic vectors $\Z$ and the other not. Therefore they applied an extension of the VEM algorithm, called the classification VEM algorithm, in which the lower bound is maximised alternatively in two steps. In the first step, the lower bound is maximised with respect to the variational distribution and the collection of the word distributions, $\P$, in the usual way of a VEM algorithm. In the second step, the maximisation is carried out in a greedy fashion with respect to the remaining latent variables, namely the latent groups in \cite{blz16}, and the latent groups and latent time clusters in \cite{cblr18}.

\subsection{Clustering approach}
For data with textual edges, while soft clustering still applies to discovering the topics of these documents, it is less clear for the nodes (users). For example, in \cite{zmlgz06} and \cite{pdbe08}, a distribution of users is associated with each group, rather than the other way round. In such cases, we only report the clustering approach for the documents in Table \ref{table.both}. \cite{scfs12} are the only ones who soft clustered both the documents and users, the latter of which can belong to multiple documents \textit{and} multiple groups. \cite{blz16} and \cite{cblr18} hard clustered the nodes and soft clustered the documents, which are the predominant approaches in SBMs and topic models, respectively.

For data with textual nodes, the models essentially assume that the same set of latent variables, that is their group memberships influence both the generation of the words and the connection between the documents. The soft clustering or mixed membership approach adopted by LDA is possible to be carried over just to a model which combines LDA with a block model, overriding the usual hard clustering approach for the latter. This is the case for all of \cite{lng09,cb10}, and \cite{hex12}.

\subsection{Number of groups/topics}
The number of groups or topics is usually fixed or determined by certain criterion, possibly because of the complexity of the model. For data with textual edges, both clustering the nodes into groups and the documents into topics are required. \cite{zmlgz06} used 6 groups and 20 topics for the Enron data set, while \cite{pdbe08} used 8 groups and 25 topics, and perplexity is used by both \cite{mwc07} and \cite{scfs12} as the criterion, the latter of which selected 10 groups and 20 topics. On the other hand, \cite{blz16} used a BIC-like criterion to find 10 groups and 5 topics, while \cite{cblr18} used the ICL to find 6 groups and 9 topics. It should be noted that, such discrepancy is due to not only the choice of the criterion but also the model itself.

For data with textual nodes, clustering the nodes into groups is equivalent to clustering the documents in topics. \cite{cb10} used various numbers (5,10,15,20,25) of topics for their data, while \cite{lng09} used perplexity as the criterion. \cite{hex12} are the only ones who used incorporated a DP in their hierarchical model, although the number of groups found is not reported. Furthermore, they fixed the hierarchical level to 2 or 3 for their data sets.

\section{Comparison} \label{sect.compare_all}
A comparison is provided in this section regarding works that incorporate longitudinal and/or topic model in SBMs and other models for graphs. Specifically, we will look at their performances on one widely studied dataset.

\subsection{Real-world example} \label{sect.compare_all_enron}
The Enron email network dataset \citep{ky04}, which contains the email communications between its employees from 1999 to 2002 before its bankcrupty in 2001, has been widely studied because of the availability of the temporal information (the time points of the email exchanges), the textual information (the actual contents of the emails), and the subsequent graph (the network of the employees). While the works reviewed have a different size of network, due to the different time periods studied, in most cases the number of nodes $n$ is around 150, except \cite{xh13} and \cite{sh19} with $n=184$. A few events happened during that period:
\begin{enumerate}
\item \label{event.enron_ceo} 2001-08-14: Then CEO Jeff Skilling resigned.
\item \label{event.enron_flaws} 2001-08-15 to 2001-08-22: The finanical flaws of the company were disclosed.
\item \label{event.enron_911} 2001-09-11: September 11 attacks, which had no direct relationship with Enron though.
\item \label{event.enron_fraud} 2001-10-31: The opening of fraud investigation by the Securities and Exchange Comission.
\item \label{event.enron_bankruptcy} 2001-12-02: The filing for bankruptcy of the company.
\end{enumerate}
These events likely affected the dynamics of the employees, which could be reflected from the email exchanges, and were looked into by different works below.

\subsubsection{Longitudinal modelling}
The dynamic SBMs reviewed in Section \ref{sect.graph_longitudinal} provided various insights into the dynamics of the network. In their dynamic MMSBM, \cite{fsx09} and \cite{xfs10} examined the temporal evolution of the mixed memberships of each employee, which can be interpreted as different (latent) roles. They found that, perhaps not surprisingly, that employees with strong connections with multiple groups and/or important positions tended to have multiple active roles most of the time. They also discovered the major changes in the mixed memberships aligned with real-life events, such as \ref{event.enron_flaws} and \ref{event.enron_bankruptcy} above. Related observations were made by \cite{xh13}, who examined the evolution of the block matrix $\C^{(t)}$ instead, and fixed the $K=7$ groups according to the known roles of the employees. They discovered that the elements of $\C^{(t)}$ spiked at event \ref{event.enron_ceo}.
  
Other applications include \cite{mrv15}, who found out that their ICL criterion did not manage to select a reasonably small $K$ under their SBM with conditional Poisson processes. Fixing $K=3$, they discovered that two groups had had intra-group communications that peaked at different periods, while the remaining group had very little activity (within itself and with other groups). While \cite{fcx15} also applied their MTI model to the Enron dataset, they only reported that it outperformed the then existing models such as the MMSBM \citep{abfx08} and the LFRM \citep{mgj09}. Similar claims of better performance over SBM were made by \cite{dbs13} in their dynamic model.

\subsubsection{Topic modelling}
Next, we look at the topics in the emails exchanged, discovered by models reviewed in Section \ref{sect.both}. \citep{mwc07} found the $K=50$ topics according to their author-recipient-topic model matched well with specific issues in the operations of a company, and such observation is echoed by the respective models of \cite{zmlgz06} and \cite{pdbe08}. Upon comparison with simply fitting an SBM to the network, they found that two employees in the same group inferred by the SBM might actually have vastly different roles in the company. This means that, while the two employees wrote to roughly the same set of employees (hence the stochastic equivalence), the difference in their roles might not be revealed without taking into account the texts of the emails. Using their community-author-recipient-topic model, \cite{pdbe08} showed that the groups found were topically meaningful, which means that different groups wrote emails on different topics, implying a successful incorporation of the textual information. However, \cite{zmlgz06,mwc07}, and\cite{pdbe08} did not compare the groups they found with the actual roles of the employees.
  
Apart from making a similar observation to \cite{pdbe08}, \cite{scfs12} managed to compute the modularity \ref{eqn.modularity}, and found that not only did their topic user community model outperformed the other models mentioned in this section across different values of $K$, it also achieved a maximimum modularity (at $K=10$) greater than 0.3, indicating significant community structure \citep{cnm04}. As modularity is not a formal criterion for model selection, \cite{scfs12} used perplexity to select an optimal number of groups, which was also around $K=10$.

In their STBM, \cite{blz16} optimised the BIC-likelihood criterion with respect to the number of groups $K$ and the number of topics \textit{simultaneously}. The optimal $K=10$ (and 5 topics) is larger than that found by an SBM alone ($K=8$). Closer investigation revealed that some employees in the same group found by SBM talked about different topics than the rest of the group, hence the splitting of a group into smaller ones found by STBM. This is in agreement with the observation by \cite{mwc07}.

\subsubsection{Longitudinal and topic modelling}
In the dynamic STBM by \cite{cblr18}, model selection was done by maximising the ICL criterion with respect to the number of groups $K$, the number of topics, \textit{and} the number of time clusters. Compared to 10 groups and 5 topics in the STBM, 6 groups, 9 topics and 4 time clusters were selected here. The 4 time clusters corresponded well (with a slight lag) to the four periods separated by events \ref{event.enron_911}, \ref{event.enron_fraud} and \ref{event.enron_bankruptcy} described at the beginning of this section. Furthermore, they found different interactions between groups at different time clusters, with different topics discussed, thus showing the need of incorporating both temporal and textual information.

\subsection{Approaches}
Similar to the comparison for SBMs in Section \ref{sect.compare_graph}, the modelling approach in topic models influences, and is influenced by other aspects discussed. For example, the use of a hierarchical process for the number of topics usually leads naturally to an MCMC algorithm, in particular a Gibbs sampler. While topic models are not the focus of this review, we related some of them with their counterparts for graphs, as similar issues regarding inference and dealing with $K$ arise in both sets of models. For a comprehensive comparison between the topic models, please see Table \ref{table.topic}.
\begin{sidewaystable}[!htbp]
  \centering
  \begin{tabular}{|l|l|l|l|l|l|}
    \hline
    \textbf{Article} & \multicolumn{2}{c|}{\textbf{Type of graph}} & \textbf{Inference} & \textbf{Clustering} & \textbf{Number of groups}\\
    \hline
    \cite{fsx09} & Binary & Directed & Variational & Soft & Criterion (BIC)\\ % paired w/ xfs10
    \cite{xfs10} & Binary & Directed & Variational & Soft & Criterion (BIC)\\ % paired w/ fsx09
    \cite{ty11,ty14} & Binary & Directed & MCMC & Hard & Dirichlet process\\
    \cite{yczgj11} & Either & Undirected & MCMC & Hard & Fixed (2,2,3)\\
    \cite{dbs13} & Valued & Directed & MCMC & Hard & Criterion (Perplexity)\\
    \cite{xh13} & Binary & Directed & Greedy algo.+SMC & Hard & Fixed (7)\\
    \cite{fcx15} & Binary & Directed & MCMC & Soft & Dirichlet process\\
    \cite{mrv15} & Binary & Undirected & Variational & Hard & Criterion (ICL)\\ % paired w/ mm17
    \cite{peixoto15a} & Either & Either & MCMC & Either & Estimated\\
    \cite{mm17} & Either & Either & Variational & Hard & Criterion (ICL)\\ % paired w/ mrv15
    \cite{zmn17} & Either & Undirected & Greedy algo. & Hard & Fixed\\
    \cite{len17} & Binary & Undirected & MCMC & Hard & Criterion (Elbow plot)\\
    \hline
    \end{tabular}
  \caption{Stochastic block models with longitudinal modelling.}
  \label{table.longitudinal}
%\end{sidewaystable}
\vspace{0.5cm}
%\begin{sidewaystable}[!htbp]
%  \centering
  \begin{tabular}{|l|l|l|l|l|}
    \hline
    \textbf{Article} & \textbf{Inference} & \textbf{Clustering} & \textbf{Number of topics} & \textbf{Longitudinal}\\
    \hline
    \cite{bnj03} & Variational & Soft & Criterion (Perplexity) & No \\
    \cite{gs04b} & MCMC & Soft & Criterion (Marginal log-like.) & No \\
    \cite{tjbb05} & MCMC & Soft & Dirichlet process & No \\
    \cite{bl06} & Variational+SMC & Soft & Fixed (20) & Yes \\
    \cite{rdc08} & MCMC & Soft & Dirichlet process & Yes \\
    \cite{pr08} & MCMC & Soft & Dirichlet process & No \\
    \cite{ax10} & MCMC & Soft & Dirichlet procees & Yes \\
    \cite{asmm10} & Unknown & Soft & Criterion (K-L divergence) & No \\
    \cite{gg11} & MCMC & Soft & Indian buffet process & No \\
    \cite{wpb11} & Variational & Soft & Dirichlet process & No \\
    \hline
    \end{tabular}
  \caption{Topic models and related clustering models.}
  \label{table.topic}
%\end{sidewaystable}
\vspace{0.5cm}
%\begin{sidewaystable}[!htbp]
%  \centering
  \begin{tabular}{|l|l|l|l|l|l|l|l|}
    \hline
    \textbf{Article} & \multicolumn{2}{c|}{\textbf{Type of graph}} & \textbf{Textual} & \textbf{Inference} & \textbf{Clustering} & \textbf{Number of groups/topics} & \textbf{Longitudinal}\\
    \hline
    \cite{zmlgz06} & Valued & directed & Edges & MCMC & Soft & Fixed (6 groups; 20 topics) & No\\
    \cite{mwc07} & Valued & directed & Edges & MCMC & Soft & Criterion (Perplexity) & No\\
    \cite{pdbe08} & Valued & directed & Edges & MCMC & Soft & Fixed (8 groups; 25 topics) & No\\
    \cite{lng09} & Binary & directed & Nodes & Variational & Soft & Criterion (Perplexity) & No\\
    \cite{cb10} & Binary & undirected & Nodes & Variational & Soft & Fixed (5,10,15,20,25 topics) & No\\
    \cite{hex12} & Binary & directed & Nodes & MCMC & Soft & Dirichlet process & No\\
    \cite{scfs12} & Valued & directed & Edges & MCMC & Soft; Soft & Criterion (Perplexity) & No\\
    \cite{blz16} & Valued & directed & Edges & Variational & Soft; Hard & Criterion (BIC-like) & No\\
    \cite{cblr18} & Valued & directed & Edges & Variational & Soft; Hard & Criterion (ICL) & Yes\\
    \hline
    \end{tabular}
  \caption{Models combining block models and topic models. Under the column ``clustering'', the first approach refers to the topics. If there is a second approach, it refers to the groups.}
  \label{table.both}
\end{sidewaystable}

\section{Discussion} \label{sect.discussion}
In this review we have seen a spectrum of statistical models for graphs, in particular the SBMs, with or without the longitudinal and/or topic modelling. They have been compared in different aspects, including the clustering approach, the inference approach, and how the number of groups $K$ is being coped with. Instead of looking at each model as a unit, we investigated them in a systematic and cross-sectional way, hopefullying presenting the landscape of the literature in a straightforward manner, so that the models most relevant to the reader can be compared with relative ease.

It should be noted that, however, these aspects are not completely independent of each other, nor are they independent of the model specification. In particular, by looking at Table \ref{table.graph}, a shift in the approach to $K$, which is highly related to the developments in the models themselves, can be seen over the years. While previous SBMs have relied on nonparametric Bayesian processes used in topic models, more recent SBMs are usually collapsed. With model parameters being integrated out, this allows model selection criteria to be computed \citep{cl15}, or tractable posterior \citep{mmfh13,nr16,peixoto17a} to be used with efficient MCMC algorithms \citep{peixoto14b}, therefore enabling $K$ to be selected or estimated without trans-dimensional methods.

Two of the recent developments are worth singling out. The first is the nested microcanonical SBM \citep{peixoto17a}, which establishes the MDL \citep{peixoto13} and its equivalence with the usual Bayesian inference approach, incorporates an efficient MCMC algorithm \citep{peixoto14a} and a hierarchical structure \citep{peixoto14b} to model $K$ and circumvent the issue with potential underfitting due to the maximum $K$ detectable in an SBM \citep{peixoto13}. One possible direction is to \replaced{}{extend these developments to MMSBMs, while retaining the scalability with the hard clustering approach. Also, it would be useful to }apply these methods and techniques to hypergraphs as well, as they are still a growing field to our knowledge.
  
The second is the increasing adopting of the DC-SBM \citep{kn11}, not only in modifying the model, but also in aiding model selection in different ways \citep{ysjkmzzz14,yan16,wb17,hqyz19}. While it is particularly useful for dealing with the degree heterogeneity, the results are more mixed when it is applied to real-world networks, as seen in Section \ref{sect.compare_graph}. Moreover, the usefulness of the DC-SBM should not be seen as an indication of the inferiority of the original SBM, as it captures different underlying structures that follow stochastic equivalence. One possible direction is to \replaced{combine the regularised SBM \citep{ls19b} and}{adopt the framework of} the weighted SBM \citep{ajc15}, and have both versions of the SBM in an unifying model. Rather than aimlessly increasing the model complexity, one should aim at modifying or extending an SBM to realistically capture the properties of real-world networks.

Future directions can also be made for extending models that combine SBMs and topic models, and can be split into the two broad categories as in Section \ref{sect.both_modelling}. For graphs with textual edges, while \cite{cblr18} currently represents the state-of-the-art among the models reviewed, a mixed membership version for the nodes could be incorporated. The number of groups/topics/time clusters could also be modelled, possibly with the associated parameters, whose dimensions grow with these numbers integrated out, potentially leading to more accurate computations of criterion and/or more efficient inference algorithms. 

For graphs with textual nodes, there are a few possible directions. One way is to develop a truly hierarchical model, potentially by incorporating a nested SBM, to infer the number of groups \textit{and} the depth of the hierarchical structure simultaneously. Another way is to combine \cite{cb10} with an MMSBM to soft cluster the nodes, accompanied by a criterion to infer the number of groups/topics. Methods by, for example, \cite{law16} can also be incorporated to obtain a scalable and efficient inference algorithm. The model by \cite{gpa18} can also be applied to account for the graph between words and documents \textit{and} the graph between documents simultaneously, in an unifying framework. Finally, the SBM by \cite{sbknm19} that allows continuous attributes on the nodes can potentially be modified to model \textit{textual} attributes instead.

\section*{List of abbreviations}
\textbf{BIC}: Bayesian information criterion;
\textbf{CTMC}: continuous-time Markov chain;
\textbf{DC}: degree-corrected;
\textbf{DP}: Dirichlet process;
\textbf{EM}: expectation-maximisation;
\textbf{EPM}: edge partition model;
\textbf{F$^2$IC}: fully factoised information criterion;
\textbf{HDP}: hierarchical Dirichlet process;
\textbf{HMC}: Hamiltonian Monte Carlo;
\textbf{ICL}: integrated complete data log-likelihood;
\textbf{ILA}: infinite latent attribute;
\textbf{IMRM}: infinite multiple relational model;
\textbf{LFRM}: latent feature relational model;
\textbf{KL}: Kullback-Leibler;
\textbf{LCA}: latent class analysis;
\textbf{LDA}: latent Dirichlet allocation;
\textbf{MCMC}: Markov chain Monte Carlo;
\textbf{MDL}: minimum description length;
\textbf{MLE}: maximum likelihood estimate;
\textbf{MMSBM}: mixed membership stochastic block model;
\textbf{MTI}: mixture time invariant;
\textbf{MTV}: mixture time variant;
\textbf{SBM}: stochastic block model;
\textbf{SMC}: sequential Monte Carlo;
\textbf{STBM}: stochastic topic block model; 
\textbf{VEM}: variational expectation-maximisation

\section*{Declarations}
\subsection*{Availability of Data and Materials}
Data sharing is not applicable to this article as no datasets were generated or analysed during the current study.

\subsection*{Competing interests}
The authors declare that they have no competing interests.

\subsection*{Funding}
This research was funded by the Engineering and Physical Sciences Research Council (EPSRC) grant DERC: Digitial Economy Research Centre (EP/M023001/1).

\subsection*{Authors' contributions}
CL compiled the articles reviewed and wrote the manuscript. Both authors reviewed and approved the final manuscript.

\subsection*{Acknowledgements}
Not applicable

\subsection*{Authors' information}
Not applicable

\bibliographystyle{agsm}
\bibliography{ref_nr}

\end{document}